\pdfoutput=1
\documentclass[journal]{IEEEtran}
\usepackage[cmex10,tbtags]{amsmath}
\usepackage{graphicx,algorithm,algpseudocode,algorithmicx,epstopdf,fixltx2e,caption,multirow,amsmath,bm,booktabs,threeparttable,cite,verbatim}
\usepackage{float}
\newfloat{algorithm}{t}{lop}
\begin{document}

\title{Comparison-based Image Quality Assessment for Parameter Selection}

\author{\IEEEauthorblockN{Haoyi Liang,}{\huge {\tiny }}
\IEEEauthorblockA{\textit{Student Member, IEEE,} and}
\and
\IEEEauthorblockN{Daniel S. Weller,}
\IEEEauthorblockA{\textit{Member, IEEE}}

\thanks{Haoyi Liang and Daniel S. Weller are with the Charles L. Brown Department of Electrical and Computer Engineering, University of Virginia, Charlottesville, VA 22904 USA (email: hl2uc@virginia.edu, dweller@virginia.edu).}
}

\maketitle
%------------------------------------------------------Abstract-------------------------------------------
\begin{abstract}
Image quality assessment (IQA) is traditionally classified into full-reference (FR) IQA and no-reference (NR) IQA according to whether the original image is required. Although NR-IQA is widely used in practical applications, room for improvement still remains because of the lack of the reference image. Inspired by the fact that in many applications, such as parameter selection, a series of distorted images are available, the authors propose a novel comparison-based image quality assessment (C-IQA) method. The new comparison-based framework parallels FR-IQA by requiring two input images, and resembles NR-IQA by not using the original image. As a result, the new comparison-based approach has more application scenarios than FR-IQA does, and takes greater advantage of the accessible information than the traditional single-input NR-IQA does. Further, C-IQA is compared with other state-of-the-art NR-IQA methods on two widely used IQA databases. Experimental results show that C-IQA outperforms the other NR-IQA methods for parameter selection, and the parameter trimming framework combined with C-IQA saves the computation of iterative image reconstruction up to 80\%.


%\textit{EDICS}---SMR-HPM, TEC-FOR
\end{abstract}

%------------------------------------------------------Introduction-------------------------------------------------
\section{Introduction}
Obtaining an image with high perceptual quality is the ultimate goal of many image processing problems, such as image reconstruction, denoising and inpaiting. However, measuring the perceptual image quality by subjective experiment is time-consuming and expensive, so designing an image quality assessment (IQA) algorithm that agrees with the human visual system (HVS)\cite{HSV_neural,HSV_IQA1,HSV_IQA2,HSV_IQA3,HSV_IQA4} is a foundational image processing objective. Moreover, most image restoration algorithms require one or more parameters to regulate the restoration process, and no-reference IQA methods can be used to guide selecting the parameters. For instance, the regularization parameter of image reconstruction\cite{parameter_trimming} is selected by a no-reference image quality index\cite{MetricQ}. However, most existing no-reference IQA algorithms output the estimated image quality based on a single distorted image, ignoring that different degraded images can provide more information together to the quality estimation of each degraded image. This observation inspires us to develop a comparison-based IQA method to fill the gap between the increasing need of parameter selection for image processing algorithms and the lack of such a NR-IQA algorithm that makes full use of the available information.

IQA algorithms are classified based on whether the reference image (the distortion-free image) is required: full-reference (FR),  reduced-reference (RR) and no-reference (NR). FR-IQA\cite{FR_IQA_review_OSU,SSIM,FR_IQA3,FR_IQA4,FR_IQA5,FR_IQA6} is a relatively well-studied area. Traditional methods like mean squared error (MSE) and signal-to-noise ratio (SNR) are used as the standard signal fidelity indices\cite{IQA_review}. A more sophisticated FR-IQA algorithm, Structural Similarity Index Method (SSIM)\cite{SSIM}, considers the structure information in images and performs well in different studies\cite{IQA_review, SSIM_cite1,SSIM_cite2,SVD_IQA}. RR-IQA algorithms\cite{RR_IQA1,RR_IQA2,RR_IQA3} require some statistical features of the reference image, such as the power spectrum, and measure the similarity of these features from the reference image and the distorted image. NR-IQA algorithms adopt two different approaches. The first kind of NR-IQA\cite{DIVIINE,BRISQUE,anisotropy,NR_IQA4,NR_IQA5,NR_IQA6,NR_IQA7} has a similar approach to the RR-IQA. The difference is that rather than extracting the features from the reference images, this kind of NR-IQA extracts statistical features from a training set. The second kind of NR-IQA algorithms\cite{MetricQ,SVD_IQA} adopts a local approach to quantifying structure as a surrogate for quality. A common implementation of the second approach calculates local scores by analyzing the quality of gradient. The overall score is synthesized by taking the average of the local scores.

Among these three kinds of IQA algorithms, speed and accuracy generally decrease from FR-IQA, RR-IQA to NR-IQA progressively. Unfortunately, reference images do not exist in many cases. Noticing that in many applications, including parameter selection and comparison of different restoration algorithms, we compare a set of distorted images with the same image content, a new comparison-based IQA (C-IQA) method is proposed in this paper. The prototype of comparison-based IQA is reflected in \cite{anisotropy}. However, the concept of comparison in \cite{anisotropy} is just implicitly mentioned by sorting the overall image qualities of a series of distorted images. In our work, the comparison-based framework is built from low level image structures, and the final output is a graph that can illustrate the local relative quality. 

After proposing the comparison-based IQA method, we demonstrate one of its applications, parameter selection. The framework of parameter trimming, first proposed in \cite{parameter_trimming}, is designed to boost the parameter selection by combining NR-IQA with parameter selection. In \cite{parameter_trimming}, parameters that do not show the potential to obtain the best result are cut during the convergence process.

The rest of the paper is organized as follows. Section \ref{NR-IQA} introduces and compares different NR-IQA methods. Section \ref{C-IQA} elaborates on the implementation details of C-IQA. The framework of parameter trimming and the technique used for image reconstruction are introduced in Section \ref{parameter_trimming}. In Section \ref{experiments} experiments are conducted on two widely used IQA databases, LIVE\cite{LIVE_dataset} and CSIQ\cite{FR_IQA_review_OSU}, to verify the accuracy of C-IQA and demonstrate the effectiveness of parameter trimming combined with C-IQA. Section \ref{conclusion} reviews the novelty and experimental results of C-IQA and suggests extensions to comparison-based IQA. 

%----------------------------------------------------------------------Previous Image Quality Assessment--------------------------------------------------
\section{Existing NR-IQA methods}  
\label{NR-IQA}
Existing NR-IQA algorithms can be classified into two types\cite{SVD_IQA}: global approaches and local approaches. The output of global approaches is a scalar number that indicates the overall quality of the image. The local approaches estimate the quality in each local patch, and an overall quality index is obtained by taking the average of the local quality indices.
%--------------------------------Scalar-output---------------------------------
\subsection{Global Approach}
The rationale behind global approaches\cite{DIVIINE,BRISQUE,anisotropy,NR_IQA4,NR_IQA5,NR_IQA6,NR_IQA7} is that the distributions of natural scene statistics (NSS) share certain common characteristics among distortion-free images, and distortions will change these characteristics. For example, it is widely-accepted that the wavelet coefficients of a natural image can be modeled by a generalized Gaussian distribution (GGD)\cite{GGD1,GGD2}.  

Because the NSS are extracted from the whole image, the final output of global approaches is a scalar number that indicates the overall quality. The advantage of global NR-IQA algorithms is that most of them are not dedicated to a specific distortion since the NSS features are a high-dimensional vector designed to be sensitive to various distortions. However, because of the high dimensionality of the statistical feature space, it is difficult to individually interpret and analyze these features quantitatively, and thus feature selection is largely an empirical work. In BRISQUE\cite{BRISQUE}, the authors treat this approach like a black box. Another drawback of the global approach is that computing these NSS features is usually time-consuming. %In our experiments, we compare our method with two state-of-art methods in this category, DIIVINE\cite{DIVIINE} and BRISQUE\cite{BRISQUE}. 
%--------------------------------graph output-------------------------------------
\subsection{Local Approach}
The output of the local approach\cite{MetricQ,anisotropy,SVD_IQA} can be a graph, which illustrates the local image quality, or a scalar number by taking the average of the graph. Because in most cases the perceived distortion varies across regions in an image, an innate advantage of the local approach is that they are able to highlight the areas where distortion is most significant. At the same time, an overall quality index is easily obtained from the local quality index. 
Because human eyes are highly sensitive to the gradient in images, and the information in images can be well represented by their gradient\cite{SSIM,MetricQ,Gradient_IQA}, the local quality index is usually evaluated using the spatial gradient information. However, the amount of the gradient, or total variation, itself is not a stable indicator of the quality\cite{SVD_IQA}. Previous works\cite{MetricQ, SVD_IQA,Structure_IQA} have shown that assessing the quality of the gradient in an image can be a promising way to evaluate the image quality. Among these works, MetricQ\cite{MetricQ} shows encouraging results choosing denoising parameters. The underlying rationale of MetricQ is that the more concentrated the gradient direction is, the better the quality of the patch is. It is a reasonable assumption since both of the two most common distortions, noise and blurring, disperse the distributions of the gradient direction. Because C-IQA makes use of the quality index defined in MetricQ, we introduce MetricQ in detail in the next paragraphs.

The local quality index used by MetricQ is based on singular values of the local gradient matrix, which have been widely used as low level features in different image processing problems, such as tracking feature selection\cite{SVD_tracking}, recognition\cite{SVD_rec} and image quality assessment \cite{SVD_IQA}. For each $n\times n$ local patch ($w$), the gradient matrix is

\begin{equation}
G =
\left[ {\begin{array}{cc}
\vdots & \vdots \\
p_x(k) & p_y(k) \\
\vdots & \vdots
\end{array} } \right],
\label{gradient_mtx}
\end{equation}

\noindent{in which $p_x(k)$ and $p_y(k)$ are the gradients of the $k^{th}$ pixel in the patch $w$ on $x$ and $y$ directions. The SVD of the gradient matrix, $G$, is defined as }

\begin{equation}
G = USV^T = U
\left[ {\begin{array}{cc}
s_1 & 0 \\
 0 & s_2
\end{array} } \right]
{\left[ {\begin{array}{cc}
V_1 & V_2 \\
\end{array} } \right]}^T,
\label{SVD_definiation}
\end{equation}
\noindent{where $U$ and $V$ are both orthonormal matrices. Vector $V_1$ is of size $2\times 1$ and corresponds to the dominant direction of the local gradient; $V_2$ is orthogonal to $V_1$ and thus represents the edge direction. Singular values, $s_1$ and $s_2$, represent the luminance variances on $V_1$ and $V_2$ respectively. Intuitively, a large $s_1$ and a small $s_2$ indicate a prominent edge in the local patch.}

In MetricQ\cite{MetricQ}, two indices reflect the quality of a local patch: Image Content Index and Coherence Index. Image Content Index is defined as
\begin{equation}
Q = s_1\frac{s_1-s_2}{s_1+s_2},
\end{equation}
\noindent{and Coherence Index is defined as}
\begin{equation}
R = \frac{s_1-s_2}{s_1+s_2}.
\label{img_content_ind}
\end{equation}

$Q$ reflects the structure prominence in a local patch and $R$ is used to determine whether a local patch is dominated by noise. % (the smaller the $R$, the more likely the local patch is dominated by noise). 
The overall score of an image is calculated by
\begin{equation}
AQ = \frac{1}{MN}\sum_{i,j:R(i,j)>\tau}Q(i,j),
\label{overall_MQ}
\end{equation}
\noindent{where $M\times N$ is the size of the image and $\tau$ is the threshold to decide whether a local patch is dominated by noise. $Q(i,j)$ and $R(i,j)$ are the Image Content Index and Coherence Index of the local patch centered at $(i,j)$ in the image. A simplified interpretation of \eqref{overall_MQ} is that $AQ$ is the average structure index of local patches that have meaningful image content.}
%\subsection{Comparison between Global and Local approaches}

%Since most computation of local NR-IQA algorithms is on the scale of the local patch, it is more feasible to provide a detailed explanation and analysis of every step in the algorithm design.
 However, because the view of the local approaches is constrained by the patch size, local approaches tend to confuse the sharpness with the blocking artifacts. %the significant artificial edges between blocks pose a serious distraction for the local algorithm to distinct them from the meaningful image content. As a result, graph-output methods tend to mistake the blocking artifacts introduced by compressive algorithm, like JPEG and JPEG2000, as meaningful structure. 
Fortunately, specialized IQA algorithms aimed at blocking artifacts can be used to evaluate this special distortion\cite{NR_IQA6,blocking_IQA}.

%Anisotropy\cite{anisotropy} is another inspiring NR-IQA. If sticking to the form of the output, Anisotropy\cite{anisotropy} should be classified into scalar-output method. However, anisotropy has most of the advantages and features of the graphic-output method, and it can be applied to evaluate the quality of small local patches, so we introduced it in this part. It is worth to point out that the framework of \cite{anisotropy} includes the idea of comparison. According to \cite{anisotropy}, for each image, the range of the local entropy peaks at the best quality image, where the peak itself is a comparative concept.

%------------------------------------------------------------------Introduction to C-IQA-----------------------------------------------------------------
\section{Comparison-based image quality assessment}
\label{C-IQA}
Previous works on IQA\cite{SSIM,anisotropy,FR_IQA5,FR_IQA7,HSV_IQA4,HSV_IQA3,HSV_IQA2,HSV_IQA1} show that IQA performance can be significantly improved by taking advantage of the characteristics of HVS. For example, the structural information that human eyes are highly sensitive to is made use by SSIM\cite{SSIM}. Traditional NR-IQA algorithms also try to exploit HVS features and make reasonable assumptions about the natural scene images, but one important aspect of HVS is ignored: comparison. In the subjective IQA experiment\cite{FR_IQA_review_OSU}, volunteers are required to evaluate the quality of an image by comparing it with a reference image, rather than giving an absolute score for the image. Although in most image processing applications, the reference image does not exist, a set of differently degraded images are available. In these cases, extending existing state-of-the-art FR-IQA algorithms to comparison-based NR-IQA algorithms is a natural thought. However, different from FR-IQA algorithms, neither of the two input image qualities is known in the comparison-based IQA framework. As a result, in a comparison-based NR-IQA algorithm, we not only measure the difference between two input images, but also assess the quality of the difference. 

\subsection{Framework of C-IQA}

\begin{figure*}[t]
	\centering{\includegraphics[width=\linewidth,trim = {0 6cm 0 3.6cm}, clip]{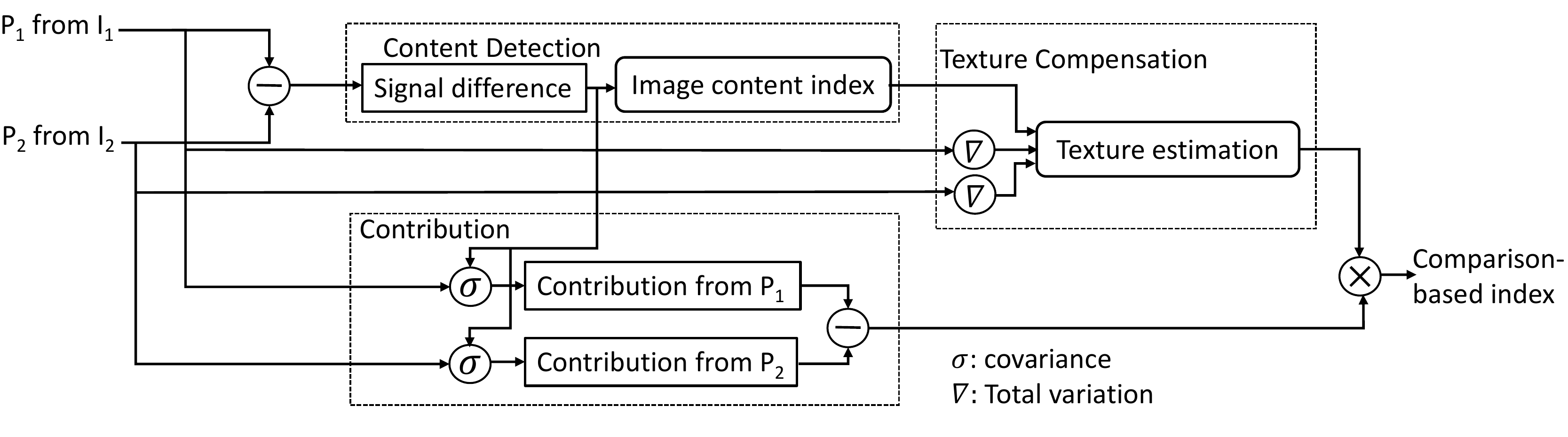}}
	\caption{Flow Chart of Comparison-based IQA: $P_1$ and $P_2$ are local patches from input images, $I_1$ and $I_2$, at the same location respectively. The Content Detection module determines whether there is a meaningful structure in the difference patch; the Contribution module calculates which patch mainly contributes to the difference patch; the Texture Compensation module compensates the distortion sensitivity difference of patches with various texture complexities. The output, comparison-based index, indicates the relative quality of $P_1$ based on $P_2$.}
	\label{C-IQA flowchat}
\end{figure*}

As shown in Fig. \ref{C-IQA flowchat}, C-IQA has two input images, $I_1$ and $I_2$, and the output indicates the relative quality of $I_1$ based on $I_2$. We refer to the second image in C-IQA as the base image to distinguish it from the reference image in FR-IQA. C-IQA consists of two basic modules: Content Detection and Contribution. The third module, Texture Compensation, is optional and its description is deferred to Section \ref{sec_textureCompensation}. In the rest of the paper, we refer to the comparison-based IQA variation composed by the two basic modules as C-IQA and the variation with three modules as CT-IQA. Content Detection determines whether the difference between two input images contains any meaningful structure, and Contribution decides which image mainly contributes to the difference. C-IQA composes these two modules by the criterion that the input image that contributes to a structured difference is better and the input image that contributes to a random difference is worse. The Texture Compensation module added in CT-IQA adjusts the distortion sensitivity difference of patches with different texture complexity\cite{SSIM,sensitivity_texture_complexity}.

\subsubsection{Content Detection}
The Content Detection module is based on the Image Content Index put forward in MetricQ\cite{MetricQ}. Different from MetricQ, this index is calculated with the difference image between two input images in C-IQA. In MetricQ, limited by the information provided by single input image, the algorithm does not know the texture complexity in the original image, and it is hard for an algorithm to tell how concentrated the gradient should be. However, by mimicking the comparative way HVS works, C-IQA removes the main image content in the images by taking the difference, and thus the Content Detection module is less influenced by the texture complexity in images.

\begin{algorithm}
\caption{Content Detection}\label{content_detection}
\begin{algorithmic}
\State{$D_p = P_1 - P_2$}
\State{$G = [d_x(D_p)\ d_y(D_p)]$} %\Comment{$G$ is a 2-column matrix composed gradients of $D_p$ on $x$ and $y$ directions
\State{$USV^T=SVD(G)$}  %\Comment{Singular Value Decomposition(SVD) of G
\State{$C_{ind} = \frac{s_1-s_2}{s_1+s_2}$} \Comment{$s_1 > s_2$}%\Comment {$S_1$ and $S_2$ are the two non-zero entries in $S$. $S_1$ is larger than $S_2$
\If{$C_{ind} > C_{thresh}$}
	\State $is\_stru = 1$  \Comment{structure}%\Comment{No significant structure is detected in $D_p$
\Else
	\State $is\_stru = -1$ \Comment{noise}%\Comment{structure is detected in $D_p$
\EndIf
\end{algorithmic}
\end{algorithm}

In Alg. \ref{content_detection}, $P_1$ and $P_2$ are two patches of size $n\times n$ from $I_1$ and $I_2$ respectively, $G$ is the same 2-column gradient matrix defined in \eqref{gradient_mtx}, $SVD(G)$ represents taking the SVD operation on $G$, and $s_1$ and $s_2$ are the singular values of $G$. $C_{thresh}$ is a constant threshold to binarize $C_{ind}$. The binary output $is\_stru$ indicates whether there is a meaningful structure in the difference of local patches.

\subsubsection{Contribution}
Once the difference is classified into noise or structure, the Contribution module is designed to find out which of the two input images mainly contributes to the difference image. In our implementation, the luminance-normalized covariance between the input image and the difference image is used to measure the contribution. 

\begin{algorithm}
\caption{Contribution}\label{contribution_alg}
\begin{algorithmic}
\State $D_p = P_1 - P_2$
\State $M_p = max(\frac{mean(P_1)+mean(P_2)}{2},\frac{1}{n\times n})$
\State $ctri1 = cov(P_1,D_p)$  
\State $ctri2 = cov(P_2,-D_p)$
\State $ctri = \frac{ctri1 - ctri2}{M_p}$
\end{algorithmic}
\end{algorithm}
In Alg. \ref{contribution_alg}, $mean(P_i)$ calculates the average of the local patch, and $cov(x_1,x_2)$ calculates the covariance between two input patches, 
\[cov(x_1,x_2) = \frac{{(x_1-mean(x_1))}^T(x_2-mean(x_2))}{n^2-1},\]
\noindent{$x_1$ and $x_2$ are vectorized patches of size $n^2\times 1$.}\\

The comparative quality index for each local patch is calculated by
\[C_Q = is\_stru \cdot ctri .\]
\noindent{The overall comparative quality of $I_1$ based on $I_2$ is}
\[CIQA(I_1,I_2) = \frac{1}{M\times N}\sum_{i,j=(n/2):(M-n/2)}{C_Q(i,j)},\] where $C_Q(i,j)$ is the local comparative quality index centered at $(i,j)$ in the image, $n\times n$ is the size of the local patch and $M\times N$ is the size of the image. Pixels that are on the margin of the image do not have $C_Q$ and thus are not included for the overall quality. A positive $CIQA(I_1,I_2)$ means $I_1$ is better than $I_2$, and the absolute value quantifies the quality difference. Due to the anti-symmetric design of the algorithm, $CIQA(I_1,I_2) = -CIQA(I_2,I_1)$.

%--------------------------------------------------Proof of C-IQA--------------------------------------------------------
\subsection{Justification of C-IQA}
\label{proof_C_IQA}

Inspired Li's work\cite{3errorSources} which claims that an IQA model should be based on three quantities: edge sharpness, random noise level and structure noise, % and Li's work\cite{3errorSources} which claims that an IQA algorithm should focus on edge sharpness, random noise and structure noise, 
we classify the distortions by residual images, the difference between a distorted image and the original image. In our new classification, distortions can be categorized into two types: introducing a random residual image, or introducing a structured residual image. In most cases, random residual images correspond to noise-like distortions and structured residual images correspond to blurring-like distortions. In this part, we prove how C-IQA works under these two distortions. 

Assume $I_{true}$ is the original image and $I_1, I_2$ are two distorted images. The residual images are calculated by,
\[e_i=I_i-I_{true},\ i = 1,2.\]
Similarly, for each patch we have
\[e_{Pi} = P_i - P_{true},\ i = 1,2.\]

\paragraph{Random residual image}
Residual images behave like noise in this case. If we assume $I_1$ is more severely distorted than $I_2$, then we have $E[{\|e_{P1}\|}_2^2]>E[{\|e_{P2}\|}_2^2]$.
The expectation of the local comparative quality index is
\begin{eqnarray}
E[C_Q]&=&E[ctri\cdot is\_stru] \nonumber \\
	  &=&E[(ctri1 - ctri2)\cdot is\_stru] \nonumber \\
	  &=&E[cov(P_1,P_1-P_2) - cov(P_2,P_2-P_1)] \nonumber\\
	  &\quad & \cdot\ E[is\_stru] \nonumber \\
	  &=&-E[cov(P_{true}+e_{P1},e_{P1}-e_{P2}) \nonumber\\
	  &\quad& -\ cov(P_{true}+e_{P2},e_{P2}-e_{P1})]\nonumber \\	  
	  &=&-E[2 \cdot cov(P_{true},e_{P1}-e_{P2}) \nonumber\\
	  &\quad& +\ cov(e_{P1},e_{P1})-cov(e_{P2},e_{P2})] \nonumber \\
	  &=&-E[cov(e_{P1},e_{P1})]+E[cov(e_{P2},e_{P2})] \nonumber \\
	  &\quad& < 0 .\nonumber
\label{def_contri}
\end{eqnarray}
The three most important properties in the derivation are the irrelevance between $P_{true}$ and $e_{Pi}$, the randomness of $e_{Pi}$, and independence of $is\_stru$ and $ctri1$, $ctri2$. The result $E[C_Q] < 0$ agrees with our assumption that $I_1$ is more severely distorted than $I_2$ and when $I_2$ is more severely distorted, the same proof shows $E[C_Q] > 0$.
%However, there is one situation where our algorithm cannot handle well. If the ``noise'' belong to the original image. As shown in Fig. \ref{snake}, noisy texture on the sands is filtered out and the local image quality increases according to our C-IQA framework. Actually this is a common phenomenon for NR-IQA algorithms, and examples for more images and NR-IQA algorithms are illustrated in Section\ref{experiments}. However, in order to correctly handle these pseudo-noise, the algorithm needs to have a good understanding of image content. For Fig. \ref{snake}, only knowing that a snake is crawling and some kinds of snakes LIVE\cite{LIVE_dataset} in the desert, can a NR-IQA tell the filtered image is better.
\paragraph{Structured residual image}
If the residual images show structured information, the most probable reason is that the image is distorted by a blurring-like distortion. Because the blurring filter acts as a low-pass filter, the residual images show a structure that is inversely related to the original image\cite{NL_denoise} to smoothen the high contrast on the edges.%(provide some pictures here). 

Without loss of generality, we assume more blurring happens in $I_1$ than $I_2$, which means $E[|e_{P1}|]>E[|e_{P2}|]$. The expectation of the local comparative quality index is
\begin{eqnarray}
E[C_Q] &=&E[ctri\cdot is\_stru] \nonumber \\
       &=&E[(ctri1 - ctri2)\cdot is\_stru] \nonumber \\
	   &=&E[cov(P_1,P_1-P_2) - cov(P_2,P_2-P_1)] \nonumber \\
	   &=&E[cov(P_{true}+e_{P1},e_{P1}-e_{P2}) \nonumber \\
	   &\quad& -\ cov(P_{true}+e_{P2},e_{P2}-e_{P1})] \nonumber \\	         
       &=& E[cov(2 \cdot P_{true},e_{P1}-e_{P2}) \nonumber\\
       &\quad& \quad +\ cov(e_{P1} + e_{P2},e_{P1}-e_{P2})] \nonumber \\
	   &=& E[cov(2 \cdot P_{true}+e_{P1}+e_{P2},e_{P1}-e_{P2}) ] \nonumber \\
	   &\quad& < 0 . \nonumber
\end{eqnarray}
The most important step in this derivation is the last step. Since  $E[|e_{P1}|]>E[|e_{P2}|]$, $e_{P1}-e_{P2}$ also demonstrates a structure that is inversely related to the original image as $e_{Pi}$. As long as the distortion is not severe enough to remove the structure in the original image, $2\cdot P_{true}+e_{P1}+e_{P2}= P_1+P_2$ is positively related to the original image. As a result, $E[cov(2 \cdot P_{true}+e_{P1}+e_{P2},e_{P1}-e_{P2}) ] < 0$, which agrees with our assumption that $I_1$ is more severely distorted than $I_2$. Following the same steps, we can show $E[C_Q] > 0$ if $I_2$ is more severe distorted than $I_1$.
%If there is a perceivable difference between $I_1$ and $I_2$, $is\_structure$ is positive. The final score $is_structure \cdot contri $ is negative, which means the $I_1$ is worse than $I_2$. But, when the difference between $I_1$ and $I_2$ is two small to be recognized as meaningful image content, wrong quality estimation will be give. We call the threshold in this phenomenon ``minimum resolution'' and more detailed discussion on this is illustrated in Sec.\ref{case_study}.

%Another kind of structure error comes from block-wise compressing algorithm, e.g., JPEG and JPEG2000. The discontinuity between neighboring patches lead to clear edges in the error map. Since the distinct features of blocking-based distortion, numerous IQA algorithms[] has been put forward on this.
\subsection{Texture Compensation}
\label{sec_textureCompensation}
We have proven that only with Content Detection and Contribution, the C-IQA can give correct results if both of the two input images are distorted by one distortion, either noise-like distortion or blurring-like distortion. However, another important property of HVS is missed in C-IQA: the response of HVS to the same distortion is texture-dependent. One example of this HVS property is that after being distorted by the same amount of Gaussian noise, the distortion in the image with simpler texture is more obvious. %, which is confirm by several wildly used IQA databases and SSIM(Some examples can be provided here). 
In this part, we first investigate such texture-based response of C-IQA and then design a compensation algorithm to adjust the sensitivity of C-IQA to different textures. We refer to the improved C-IQA as CT-IQA.

\begin{figure}[t]
	\begin{minipage}[t]{.97\linewidth}
		\begin{minipage}[t]{.24\linewidth}
			\includegraphics[width= 0.99\linewidth]{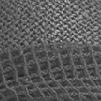}	
		\end{minipage}	
		\begin{minipage}[t]{0.24\linewidth}		
			\includegraphics[width= 0.99\linewidth]{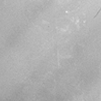}	
		\end{minipage}
		\begin{minipage}[t]{0.24\linewidth}	
			\includegraphics[width= 0.99\linewidth]{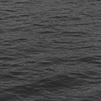}		
		\end{minipage}
		\begin{minipage}[t]{0.24\linewidth}	
			\includegraphics[width= 0.99\linewidth]{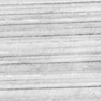}		
		\end{minipage}
	\end{minipage}	
	\begin{minipage}[t]{.97\linewidth}
		\begin{minipage}[t]{.24\linewidth}
			\includegraphics[width= 0.99\linewidth]{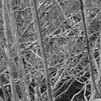}		
		\end{minipage}	
		\begin{minipage}[t]{0.24\linewidth}		
			\includegraphics[width= 0.99\linewidth]{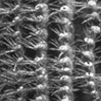}			
		\end{minipage}
		\begin{minipage}[t]{0.24\linewidth}		
			\includegraphics[width= 0.99\linewidth]{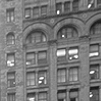}		
		\end{minipage}
		\begin{minipage}[t]{0.24\linewidth}		
			\includegraphics[width= 0.99\linewidth]{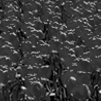}		
		\end{minipage}
	\end{minipage}	
	\caption{Patch samples are selected from LIVE\cite{LIVE_dataset} and CSIQ\cite{FR_IQA_review_OSU} to verify the texture compensation in C-IQA.}
	\label{patch_sample}
\end{figure}

\begin{figure}[t]
	\begin{minipage}[t]{.49\linewidth}
		\centering
		\includegraphics[width=4.0cm]{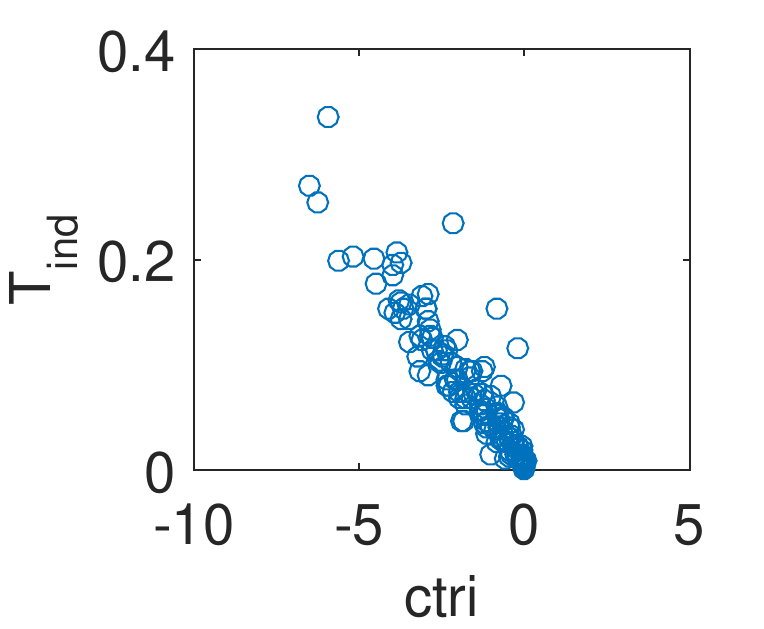}
		(a) Blurring ($\rho=-0.92$)
	\end{minipage}	
	\begin{minipage}[t]{0.49\linewidth}
		\centering
		\includegraphics[width=4.0cm]{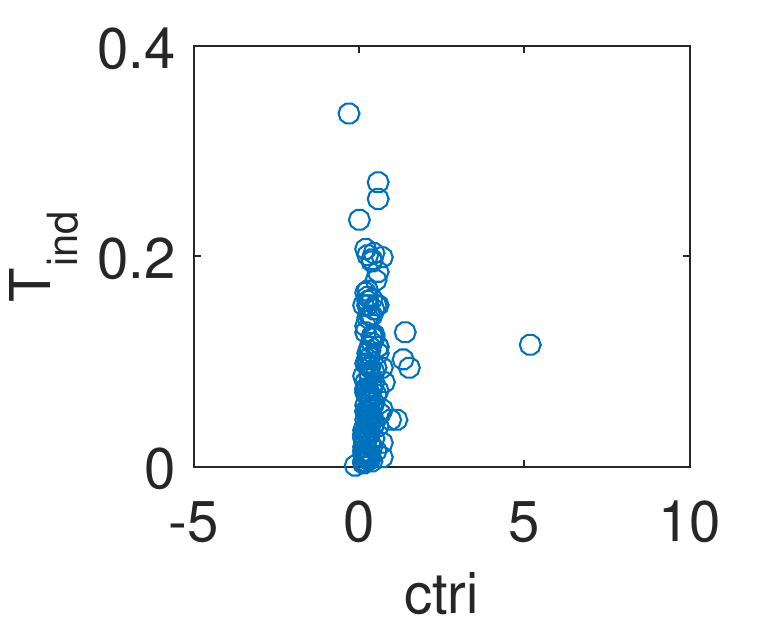}
		(b) Noise ($\rho=0.10$)
	\end{minipage}
	\caption{Relations between $ctri$ and $T\_ind$. Each circle in the figure represents a sample patch. All the sample patches are degraded by the same amount of distortion for blurring and noise.}
	\label{contri}
\end{figure}

In C-IQA, Content Detection is a qualitative module that detects the meaningful structure and the Contribution module quantifies the relative quality. Therefore, the Contribution module may implicitly include compensation. We design an experiment to explore the relation between the texture complexity and the output of Contribution, $ctri$. In this experiment, 140 patches of size $101\times 101$ with homogeneous texture are selected from LIVE\cite{LIVE_dataset} and CSIQ\cite{FR_IQA_review_OSU}, and six samples of these patches are shown in Fig. \ref{patch_sample}. As the representatives of blurring-like and noise-like distortions, bilateral filter and Gaussian noise with the same parameters are applied to each patch. According to the Weber-Fechner law\cite{weber}, we use luminance-normalized total variation as the perceived texture complexity, $T\_ind = \frac{TV(P)}{mean(P)}$, where $TV(P)$ is the total variation in the original patch and $mean(P)$ is the average of the original patch. The relation between $T\_ind$ and $ctri$ are plotted in Fig. \ref{contri}, in which each circle represents a patch. From Fig. \ref{contri}, it is clear that $ctri$ is almost linear related to texture complexity, $T\_ind$, when blurring happens. On the contrary, $T\_ind$ shows no relation with $ctri$ when the distortion is noise. The reason for this is that blurring is a highly image-dependent distortion, and the residual image is more prominent at areas where total variation is high. After figuring out the blurring sensitivity compensation mechanism in C-IQA, we need to design an algorithm to compensate the sensitivity difference to noise.

\begin{algorithm}	
	\caption{Texture Compensation}	
	\label{Textureness_Compensation}
	\begin{algorithmic}
		\State{$T1\_ind = \frac{TV(P_1)}{mean(P_1)}$} %\Comment{$TV(P_1)$ is the total variance in $P_1$}
		\State{$T2\_ind = \frac{TV(P_2)}{mean(P_2)}$} %\Comment{$TV(P_2)$ is the total variance in $P_2$}
		\If{is\_stru = 1}
		\State$T\_ind = max\{T1\_ind,T2\_ind\}$;	
		\Else
		\State$T\_ind = min\{T1\_ind,T2\_ind\}$;	
		\EndIf
		\State$S\_ind = log(1+\frac{1}{C_1\times T\_ind})$ %\Comment{transfer texture index to smooth index}
		
		\If{is\_stru = 1} 
		\State$weight =1 $  %\Comment{No compensation of structured distortion}
		\Else
		\State $weight = -S\_ind$ %\Comment{compensate noisy distortion according to the texture complexity}
		\EndIf	
	\end{algorithmic}
\end{algorithm}

Because noise-like distortion tends to increase the total variation while blurring-like distortion tends to decrease the total variation, Alg. \ref{Textureness_Compensation} uses the output of Content Detection to synthesize $T1\_ind$ and $T2\_ind$ into $T\_ind$. After texture complexity estimation, we transfer $T\_ind$ to the smoothness index, $S_{ind}$, and compensate the sensitivity to noise.

In CT-IQA, the comparative quality index for each local patch is \[CT_Q = is\_stru \cdot ctri \cdot weight.\] The overall comparative quality of $I_1$ based on $I_2$ is calculated by taking the average of local comparative quality index as C-IQA does.
\subsection{Comparison between CT-IQA and SSIM} 
SSIM consists of three components: structure (loss of correlation), luminance (mean distortion) and contrast (variance distortion). In CT-IQA, the outputs of Content Detection and Texture Compensation provide a ``reference image'' (the difference image) and the quality of the ``reference image''. The luminance and the contrast of an input image together determine the contribution of the input image to the ``reference image''. Therefore, Content Detection and Texture Compensation of CT-IQA together play the role of the structure part in SSIM. The difference is that without knowing which image has the better quality, CT-IQA has to analyze the quality of the structure in the ``reference image'', rather than only measuring the structure distance as SSIM does. The Contribution module in CT-IQA is similar to the functions of luminance and contrast parts together in SSIM.  %Similar to SSIM, to avoid the unstable results, we also add a terms to the normalization term (luminance). 

%--------------------------------------------------------------------------Application of Parameter Selection-----------------------------------------------
\section{Parameter Selection}
\label{parameter_trimming}
As the motivation of C-IQA mentioned in the introduction, most image processing algorithms contain user-defined parameters (these image processing algorithms are referred as ``target algorithms'' in the following to differ from IQA algorithms). Parameter selection \cite{parameter_select1,parameter_select2,parameter_select3,parameter_select4,parameter_select5,parameter_select6,parameter_select7,MetricQ,parameter_trimming} is of importance to these target algorithms. By parameter selection, some of these target algorithms\cite{parameter_select5,parameter_select6} achieve a faster convergence rate; some \cite{parameter_select3,parameter_select4} obtain a better restored image.

A traditional approach to parameter selection\cite{parameter_select1,parameter_select2,parameter_select3,parameter_select4} is selecting the parameters after the convergence of all the target algorithm instances with a perceptual quality monitor, usually a NR-IQA algorithm. However, since either the target algorithms converge quickly \cite{parameter_select4, MetricQ} or the NR-IQA algorithm is time-consuming \cite{parameter_select3}, computational efficiency is not considered in previous works. For instance, the denoising parameter selection in \cite{MetricQ} involves experiments with 30 parameter candidates and 20 iterations/candidate. In situations where target algorithms converge slowly or the set of parameter candidates is large, assessing image qualities and selecting the best parameter after all the algorithm instances converge would be too time-consuming to be practical.
Instead of placing the quality monitor at the output end, a novel parameter trimming framework proposed in \cite{parameter_trimming} integrates the quality monitor into the target algorithms. By doing so, parameters that do not have the potential to achieve good results are trimmed before convergence. In this section, we use image reconstruction as the application to illustrate the parameter trimming framework because a regularized iterative algorithm is usually adopted to obtain superior reconstructed results.

\subsection{Image Reconstruction}
\label{image_reconstruction}
Total variation (TV) reconstruction\cite{TV_recon} is aimed at minimizing the cost function,
\begin{equation}
\label{recon_target}
E_\beta(x)=\beta{\|Dx\|}_1+\frac{1}{2}{\|Sx-y\|}_2^2\ ,
\end{equation}
where $x$ is the reconstructed image, $y$ is the observed incomplete data set, $S$ is the system matrix, $D$ represents the difference matrix, and the TV regularizer ${\|Dx\|}_1$ combines gradients on two directions isotropically. In our implementation, $S = R\mathcal{F}$, where $R$ represents the subsampling matrix and $\mathcal{F}$ represents the Fourier transform matrix. The regularization parameter $\beta$ controls the sharpness of the reconstructed result. Large $\beta$ will oversmooth the reconstructed image, while small $\beta$ will leave residual noise. A proper $\beta$ is crucial to the performance of TV reconstruction. Split Bregman iteration\cite{split_bregman} is used to solve \eqref{recon_target}. By making the replacement $d \leftarrow Dx$ and introducing the dual variable $b$, the split formulation of \eqref{recon_target} becomes:
\begin{equation}
\label{recon_sub} 
\begin{split}
\min_{x,d}\beta \|d\|_1 + \frac{1}{2}\|Sy&-y\|_2^2 + \frac{\mu}{2}\|d-Dx-b\|_2^2\ ,\\
&s.t.\ d=Dx.
\end{split}
\end{equation}

The Split Bregman iteration solution to \eqref{recon_sub} is Alg. \ref{Split_Bregman}.
In Alg. \ref{Split_Bregman} we use the notation $K=(R^TR-\mu\mathcal{F}D^TD\mathcal{F}^{-1})$, $L_k=(\mathcal{F}^TR^Ty+\mu D^T (d^k-b^k ))$ and  $s^k=\sqrt{{|Dx^k+b^k|}^2 }$. $\mu$ is set as $0.01\beta$ to ensure a fast convergence rate.
\begin{algorithm}
\caption{Split Bregman}\label{Split_Bregman}
\begin{algorithmic}
\State{Initialize: $x^0=0,d^0=b^0=0$}
\While{stop criterion is not satisfied}
\[x^{k+1} = \mathcal{F}^{-1}K^{-1}L_k\]
\[d_{k+1}=\max(s^k-\frac{1}{\mu},0)\frac{Dx^k+b^k}{s^k}\]
\[b^{k+1}=b^k+(Dx^k-d^{k+1})\]
\EndWhile
\end{algorithmic}
\end{algorithm}

% $\mu$ is a Split Bregman penalty parameter, which controls the convergence rate and is set to 0.01 in our implementation. %It is worth to point out that since $D$ is circulant, $\mathcal{F}D^TD\mathcal{F}^{-1}$ is diagonal.}

\begin{figure}[t]
	\begin{minipage}[t]{0.24\linewidth}
		\centering{\includegraphics[width= 0.99\linewidth]{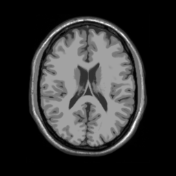}}	
		(a)	
	\end{minipage}
	\begin{minipage}[t]{0.24\linewidth}
		\centering{\includegraphics[width=0.99\linewidth]{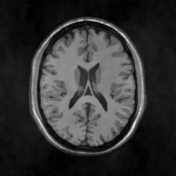}}
		(b)
	\end{minipage}
	\begin{minipage}[t]{0.24\linewidth}
		\centering{\includegraphics[width=0.99\linewidth]{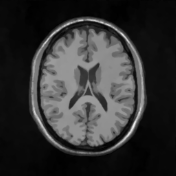}}
		(c)
	\end{minipage}
	\begin{minipage}[t]{0.24\linewidth}
		\centering{\includegraphics[width=0.99\linewidth]{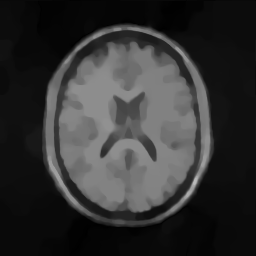}}
		(d)
	\end{minipage}
	\caption{(a): original Brain image\cite{Brain}; (b): reconstructed result with $\beta = 1.22\times 10^{-6}$; (c): reconstruction result with  $\beta = 4.46\times 10^{-1}$; (d): reconstructed result with  $\beta = 10$.}
	\label{recon_examples}
\end{figure}

\begin{figure}[t]
\begin{minipage}[t]{0.48\linewidth}
\centering{\includegraphics[width=5cm,trim = {0.5cm 0 0.5cm 0.5cm},clip]{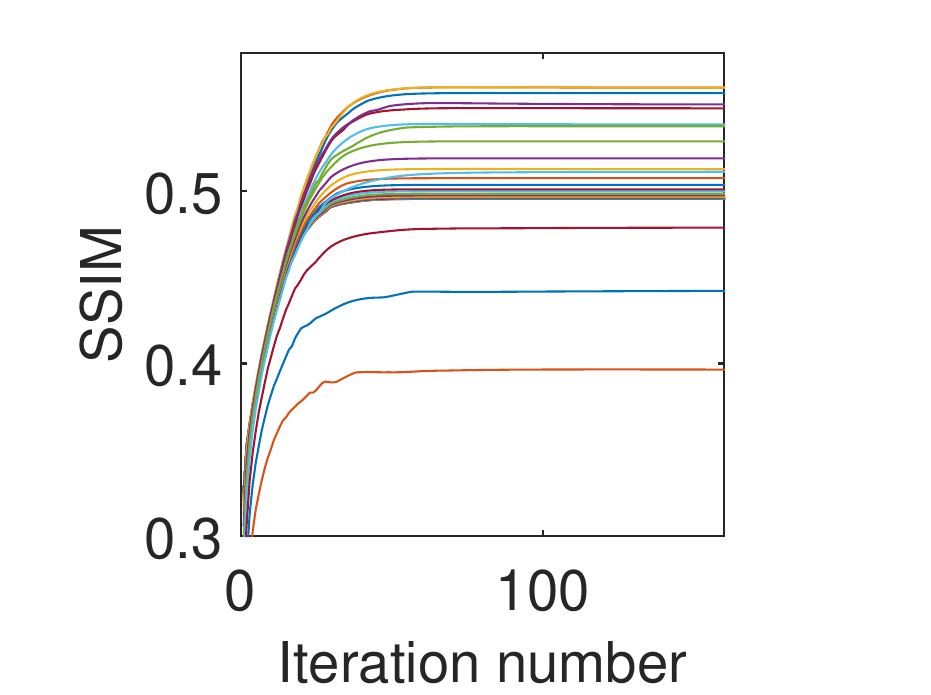}}
(a)
\end{minipage}
\begin{minipage}[t]{0.28\linewidth}
\centering{\includegraphics[width=5cm,trim = {0.4cm 0  0.5cm 0.5cm},clip]{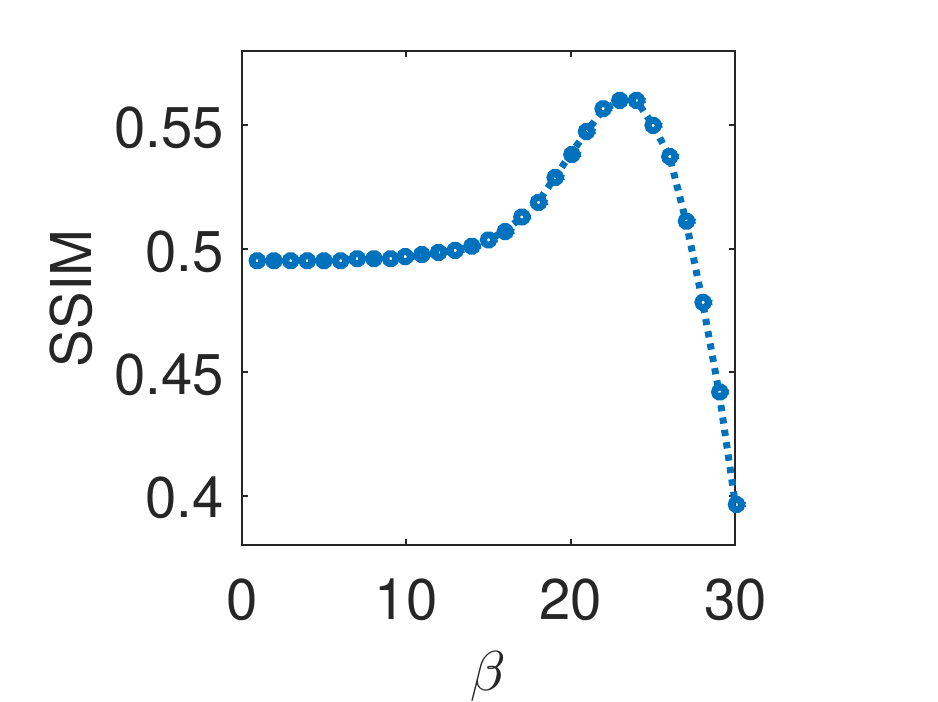}}
\centering{(b)}
\end{minipage}
\caption{(a): Each line corresponds to an algorithm instance with a parameter candidate. (b): Reconstructed result qualities of different parameter candidates after 160 iterations.}
\label{recon_converge_line}
\end{figure}

To illustrate the necessity of parameter selection of TV reconstruction, the Brain image\cite{Brain} is reconstructed with 30 values of  $\beta$. These candidate values of $\beta$ are uniformly sampled from $1.22 \times 10^{-6}$ to $10$ in logarithmic scale and three of the reconstructed results are shown in Fig. \ref{recon_examples}.
The image quality indices during the convergence process are plotted in Fig. \ref{recon_converge_line}(a), and each line corresponds to one parameter candidate. The final reconstructed image qualities are plotted in Fig. \ref{recon_converge_line}(b). 

\subsection{Parameter trimming}
In \cite{parameter_trimming}, we first proposed a parameter trimming framework that combines the quality index with target algorithms to carry out the parameter selection before convergence. Assume $I_m^i$ is the reconstructed result of the $m^{th}$ parameter candidate at the $i^{th}$ iteration. The trimming decision is made based on three indices, $s_m^i$, $g_m^i$ and $p_m^i$, which are the reconstructed quality, the quality increasing gradient and the prediction of the quality of $I_m^i$ respectively. Because the image quality index we use here is a comparison-based index, the definitions of the these three indices are modified to fit CT-IQA into the parameter trimming framework in \cite{parameter_trimming}. Denoting the best reconstructed result at the $i^{th}$ iteration is $best_i$, it satisfies $CTIQA(I_{best_i}^i,I_{best_i-1}^i) \ge 0$ and $CTIQA(I_{best_i}^i,I_{best_i+1}^i) \ge 0$. The three indices used for parameter trimming, $s_m^i$, $g_m^i$ and $p_m^i$, are defined as,
\[s_m^i = CTIQA(I_m^i,I_{best_i}^i),\]
\[g_m^i = CTIQA(I_m^i,I_{best_{i-1}}^{i-1}) - CTIQA(I_m^{i-1},I_{best_{i-1}}^{i-1}),\]
\[p_m^i = s_m^i + pre_{len}\cdot g_m^i.\]
We set $pre_{len} = 4$ in all the experiments. More examples of the reconstruction process and the changing of these three indices during the trimming process are shown in Section \ref{experiments}.
%----------------------------------------------------------------Experiments-------------------------------------------------
\section{Experiments}
\label{experiments}

\begin{figure}[t]
\begin{minipage}[t]{0.48\linewidth}
\centering
\includegraphics[height = 3.2cm]{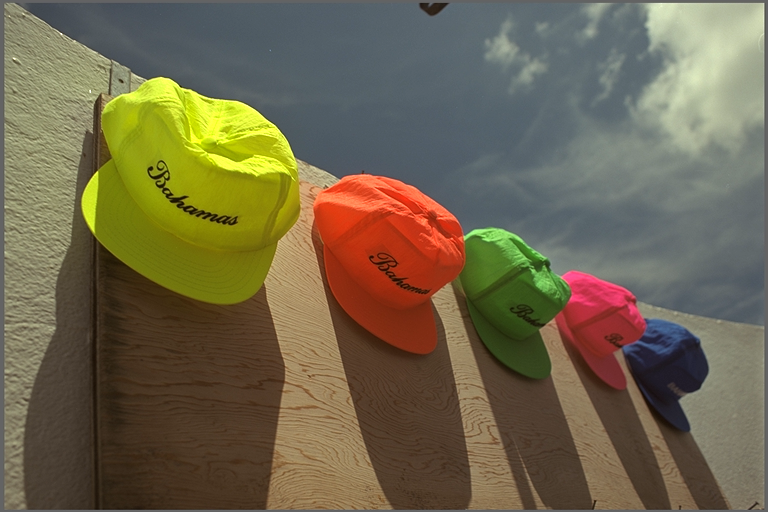}
(a) ``caps''
\label{sample_images_case_study_caps}
\end{minipage}
\begin{minipage}[t]{0.48\linewidth}
\centering
\includegraphics[height = 3.2cm]{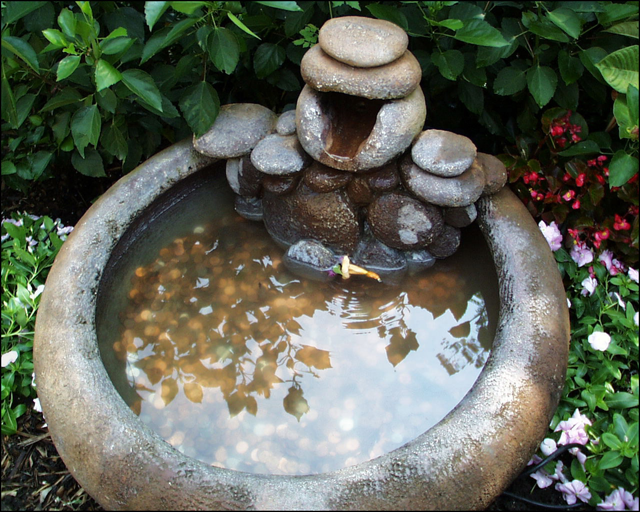}
(b) ``coinsinfountain''
\end{minipage}
\caption{Images from LIVE\cite{LIVE_dataset} used for case study in Section \ref{case_study}}
\label{sample_images_case_study}
\end{figure}

We first introduce two key properties (consistency and minimum resolution) of C-IQA/CT-IQA in Section \ref{case_study}. In the next two parts, more comprehensive experiments on two databases are conducted to verify the effectiveness of C-IQA/CT-IQA and their applications to parameter selection.

The other NR-IQA algorithms that we use to compare C-IQA/CT-IQA with are DIIVINE (DII)\cite{DIVIINE}, BRISQUE (BRI)\cite{BRISQUE}, MetricQ (MQ)\cite{MetricQ} and Anisotropy (Ani)\cite{anisotropy}. A widely accepted FR-IQA algorithm, SSIM\cite{SSIM}, is used as the ground truth to evaluate the performance of different NR-IQA algorithms. Two IQA databases used in the experiments are LIVE\cite{LIVE_dataset} and CISQ\cite{FR_IQA_review_OSU}. Parameters in C-IQA/CT-IQA are set as $C_{thresh}=0.12$, $C_1 = 4.6$ and the size of local patch is $9\times 9$.
%--------------------------------------------------Analysis on several images-----------------------
\subsection{Case study to explore two key properties of C-IQA/CT-IQA}
\label{case_study}
\begin{figure}[t]
\begin{minipage}[t]{0.99\linewidth}
\centering
\includegraphics[width = 0.99\linewidth,trim={7cm 0cm 18cm 0cm},clip]{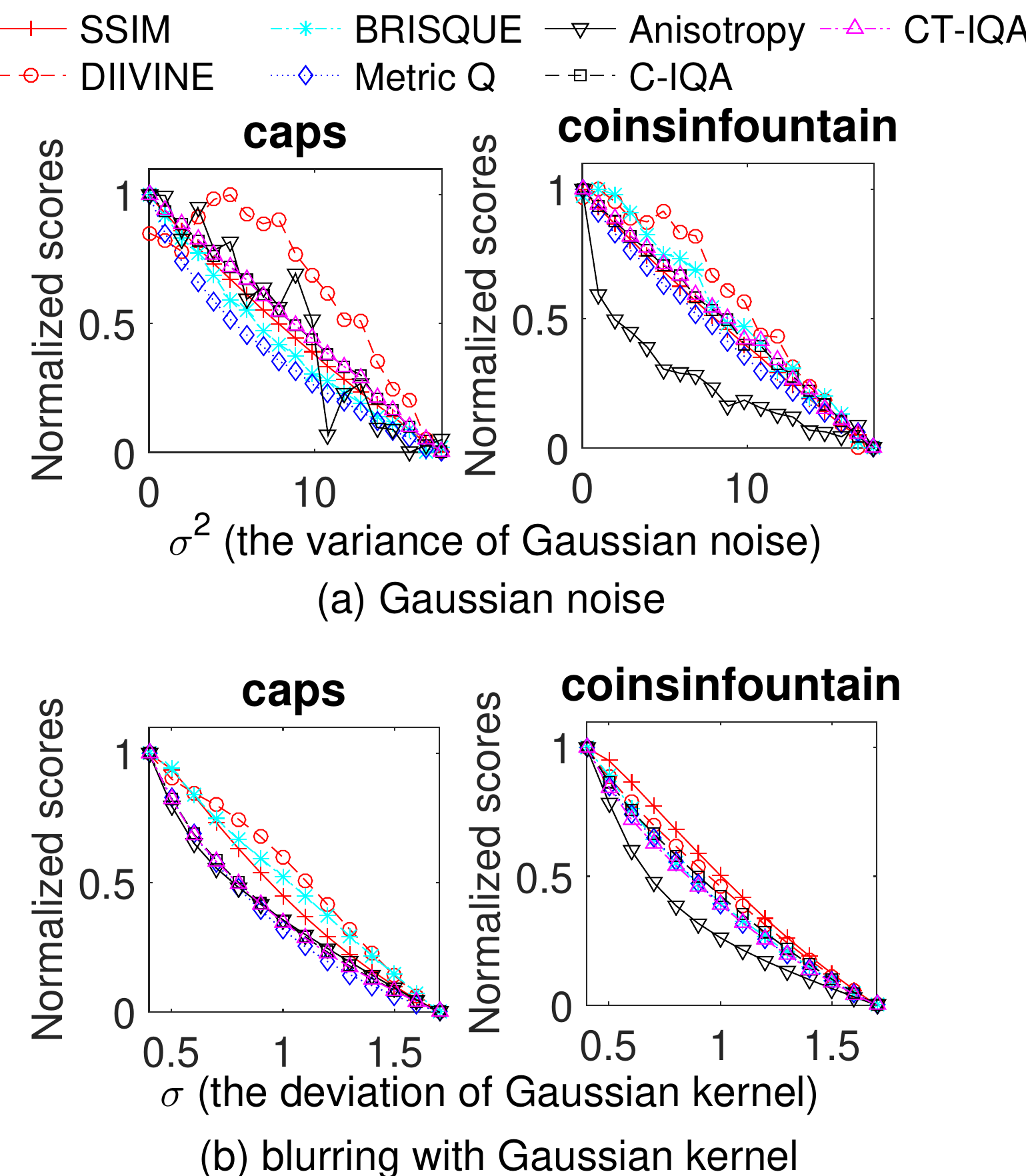}
\centering
\end{minipage}
\caption{Consistency of different IQA algorithms on noise and blurring. The scores of Anisotropy and DIIVINE are inconsistent on Gaussian noise.}
\label{consistence_fig}
\end{figure}

\begin{figure*}[t]
\begin{minipage}[t]{0.32\linewidth}
\centering
\includegraphics[width=0.99\linewidth,trim ={7cm 0cm 0.5cm 4cm},clip]{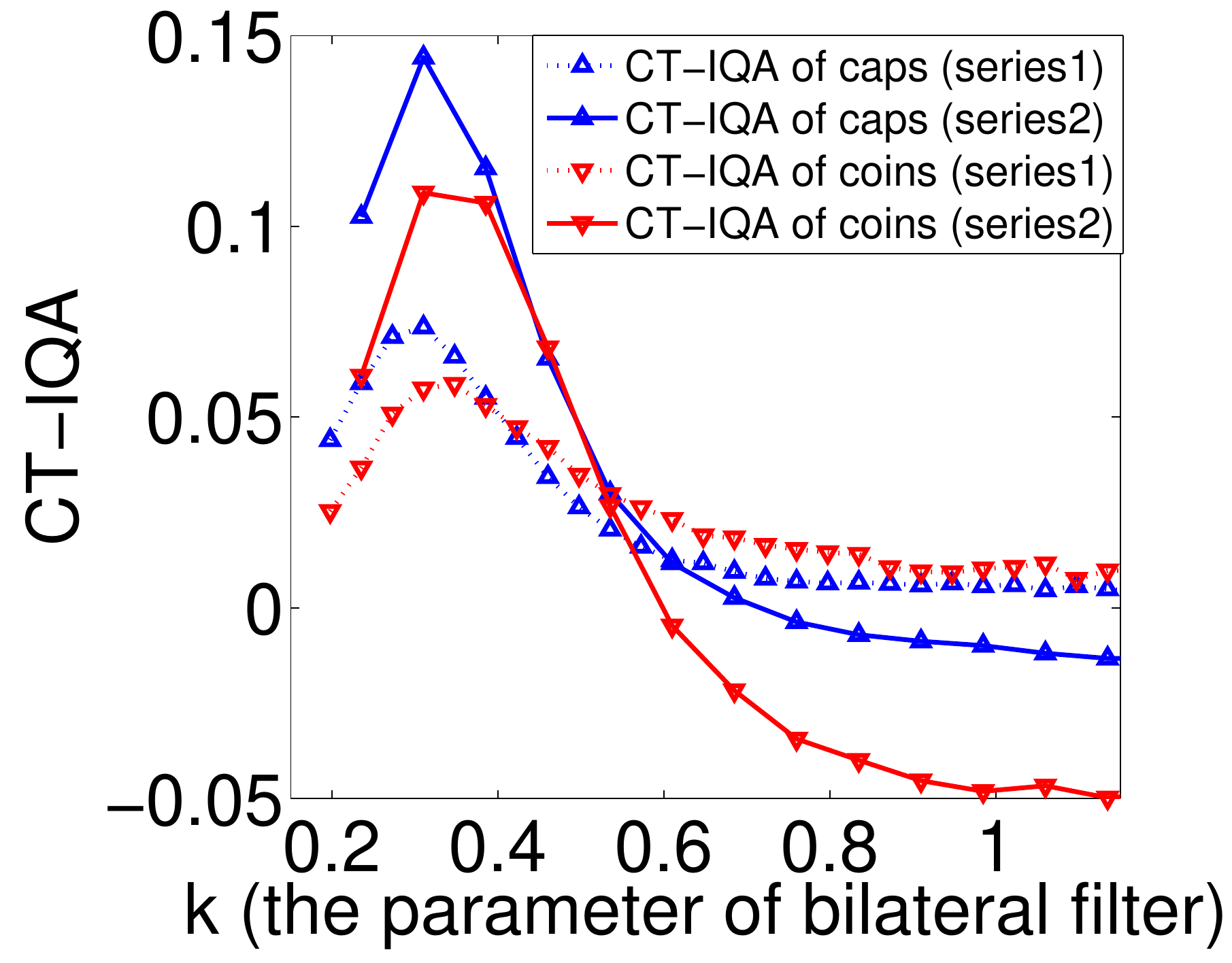}
(a) 
\end{minipage}
\begin{minipage}[t]{0.32\linewidth}
\centering
\includegraphics[width=0.99\linewidth,trim ={7.5cm 0cm 0cm 3cm},clip]{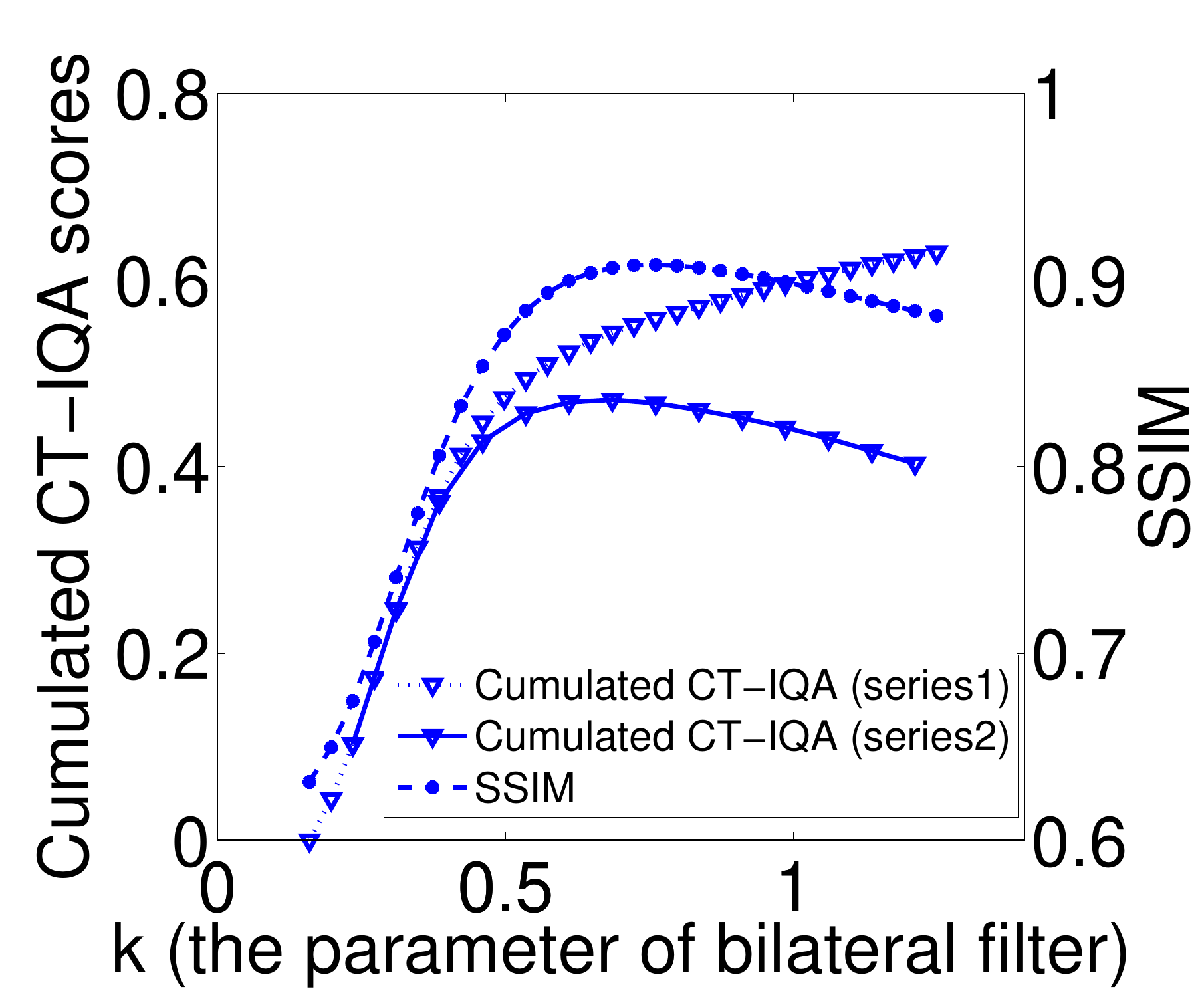}
(b) 
\end{minipage}
\begin{minipage}[t]{0.32\linewidth}
\centering
\includegraphics[width=0.99\linewidth,trim ={7.5cm 0cm 0cm 3cm},clip]{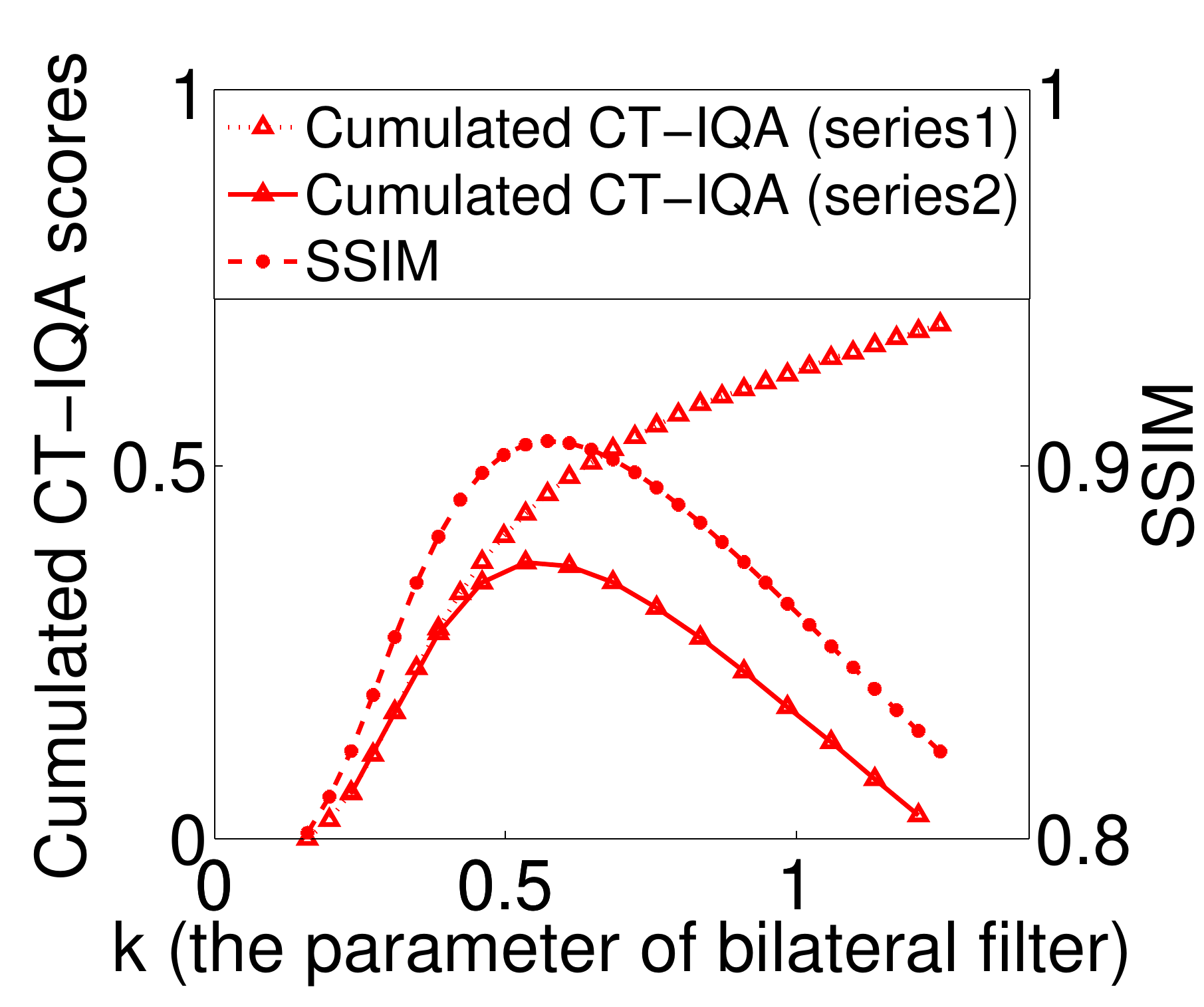}
(b)
\end{minipage}
\caption{Minimum resolution of Comparison-based IQA algorithm: (a) CT-IQA scores of denoised images compared with their previous images (series1) and the one before previous images (series2); (b) Accumulated CT-IQA scores in (a) and SSIM scores of ``caps''; (c) Accumulated CT-IQA scores in (a) and SSIM scores of ``coinsinfountain''}
\label{minimum_resolution}
\end{figure*}

Since the comparison-based IQA is a brand-new approach, some new properties arise. In this section, we illustrate these properties and corresponding solutions based on two images from LIVE\cite{LIVE_dataset} as shown in Fig. \ref{sample_images_case_study}.

%--------------------------------consistence of distortion on single image-----------
\subsubsection{Consistency on single kind of distortion}
\label{consistence_part}
Consistency on a single kind of distortion is one of the basic requirements of an IQA algorithm. Because distortions can be generally classified into two categories\cite{denoise_review}, noise and blur, we check the consistency of different NR-IQA algorithms on Gaussian noise and blurring with the Gaussian kernel respectively. 
%(Another reason we choose these two two kinds of distortion is that inappropriate parameters for image reconstruction and most image enhancing problems basically introduce noise or blur  the results.)

For each distortion, a series of increasingly distorted images, whose SSIM indices uniformly range from 0.85 to 1, are evaluated by different NR-IQA algorithms. Since we are interested in the trend of each IQA algorithm, scores obtained from different IQA algorithms are normalized between [0, 1] and the original images are used as the base images for C-IQA/CT-IQA. It is clear from Fig. \ref{consistence_fig} that all of NR-IQA algorithms produce consistent results for blurring, but in the noise case, DIIVINE and Anisotropy are inconsistence. %A possible explanation to this is that unlike blurring, noise does not degrade images in a continuously increasing way, leading to the advanced statistic features which global approaches are based on do not change smoothly.
%thus vulnerable to this\cite{foolDeepLearning}(should find more clear and accurate words to describe this...).

%-------------------------------------------------Minimum resolution----------------------------
\subsubsection{Minimum resolution}
\label{minimum_resolution_part}
Similar to HVS, IQA algorithms are not able to make a convincing quality comparison between images whose difference is sufficiently small.
%(maybe the easiest way to interpret tiny is large SSIM score between two images). 
In this part, we define the minimum mean squared difference (MSD) between two images required to make a convincing quality comparison as the minimum resolution. It is worth noticing that minimum resolutions vary over different distortions and different IQA algorithms.

For the traditional single-image-input NR-IQA algorithms, minimum resolutions can be regarded as the minimum MSD required to ensure consistency on a series of increasingly distorted images. The unwanted fluctuations of DIIVINE and Anisotropy in Fig. \ref{consistence_fig}(a) indicate that the MSD between the adjacent images is less than their minimum resolutions of Gaussian noise.

However, under the comparison-based framework, a distorted image has different scores compared with different base images. We cannot refer to the consistency to define the minimum resolution for a comparison-based IQA algorithm. The minimum resolution for comparison-based IQA is defined as the minimum MSD required to preserve transitivity among a series of distorted images. We conduct an experiment on the images in Fig. \ref{sample_images_case_study} to demonstrate the transitivity. Assume $I_{org}$ is the original image, and $I_1$ is created by adding Gaussian noise to $I_{org}$. A series of gradually filtered images, $(I_1,I_2,\cdots,I_N)$, are denoised by bilateral filters $BF_{(r,d)}$, where $r$ and $d$ are the variances of Gaussian range kernel for smoothing differences in intensities and Guassian spatial kernel for smoothing differences in coordinates. For simplicity, we reduce the parameters of bilateral filters to one by fixing the ratio between $r$ and $d$, $BF_{k} = BF_{(0.1k,3k)}$. In Fig. \ref{minimum_resolution}(a), we show the CT-IQA scores of each image compared with its previous one in the denoised sequence ($series1$) and the CT-IQA scores compared with the one before its previous one ($series2$). We can see that CT-IQA scores in $series1$ are always positive, but pass $0$ in $series2$, which means the denoised image qualities are monotonically increasing in $series1$, but reach a peak in $series2$. In Fig. \ref{minimum_resolution}(b) and Fig. \ref{minimum_resolution}(c), we plot the cumulated CT-IQA scores in $series1$ and $series2$. It is clear that CT-IQA fails to characterize the trend of image quality in $series1$, but successfully reflects the peak in $series2$. In this example, the MSD between adjacent images in $series1$ is below the minimum resolution of the bilateral filter, but the MSD between adjacent images in $series2$ is above the minimum resolution of the bilateral filter. 
There are two ways to avoid the unwanted result of operating below minimum resolution. First, increase the difference between adjacent images by increasing the parameter steps. Second, avoid comparing the adjacent images in a series of increasingly distorted images. The Key Image algorithm introduced in the next part is an implementation of the second way.

%----------------------------------------------Experiments on datasets-----------------------------

\subsection{Experiment verification on databases}
%Since distortion can be generally classified into noise and blurring, balance ability between blurring and noise in an image is an important capability of IQA algorithms. 
In this part, we evaluate the performance of different NR-IQA algorithms by comparing their balance abilities between noise and blurring on two databases. 

\subsubsection{Balance ability among different distortions}
\begin{figure}	
	\includegraphics[width = 0.99\linewidth,trim = {2.6cm 0 2.6cm 0}, clip]{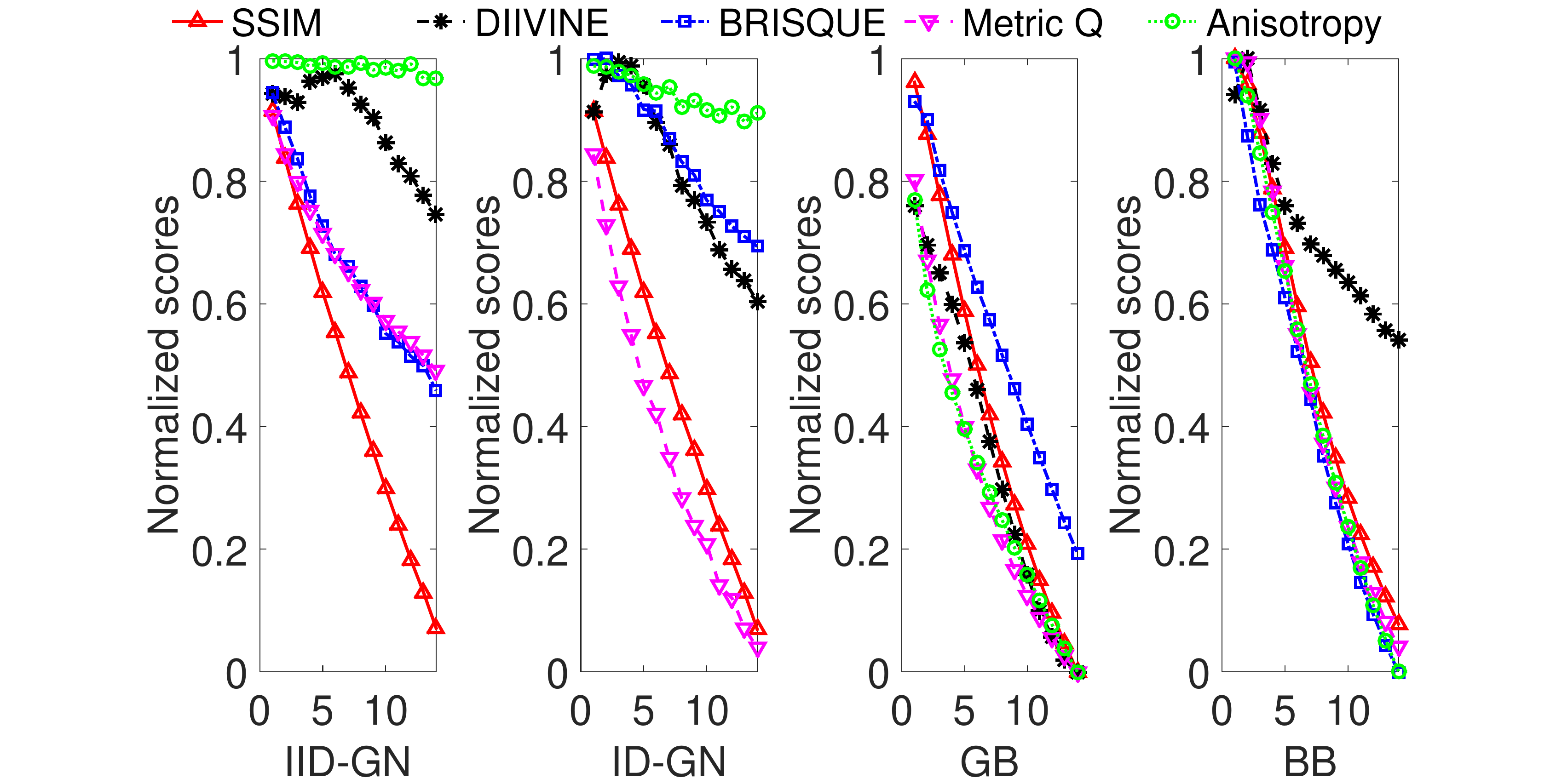}
	\caption{IQAs scores of 60 degraded ``caps'' by four distortions}
	\label{blanace_separately_example}
\end{figure}

In this experiment, four distortions are applied to each original image and four series of increasingly distorted images are created: independent and identically distributed Gaussian noise (IID-GN), Zero-mean Gaussian noise with an intensity-dependent variance (ID-GN), blurring with Gaussian kernel (GB) and blurring with bilateral filter (BB). We reduce the parameters of the bilateral filter to one parameter the same way as we did in Section \ref{case_study}. For each image under each kind of distortion, we first search the distortion parameter to ensure the SSIM index of the distorted image is between $0.85\pm 0.01$, and then uniformly sample the other 14 parameters between 0 and the searched distortion parameter. In Fig. \ref{blanace_separately_example}, we show the IQA scores of 60 distorted ``caps'' by different IQA algorithms. All the scores are also normalized to $[0,1]$ for each IQA algorithm.

Since it is justified for each IQA algorithm to have its own sensitivity properties at different distortion levels, we evaluate the balance ability among different distortions of each NR-IQA algorithm at 14 distortion levels. Eight distorted images with adjacent distortion parameters of four distortions are combined into an image set. The average SSIM index of all the eight images is the distortion level of this set. Therefore, we have 14 sets of distorted images and rank the eight distorted images in each set according to different IQA algorithms. The weighted inversion numbers\cite{inversion_number} between the ranking results by NR-IQA algorithms and by SSIM are used to evaluate the performance of different NR-IQA algorithms. Assume $(I_1,\cdots,I_N)$ is the ranking sequence according to a NR-IQA algorithm from low quality to high quality, the weighted inversion number in our experiment is defined as
\[WInv_{num}=\sum_{i=1:N}{\sum_{j=i+1:N}{max(0,SSIM(I_i)-SSIM(I_j))}}.\]

%It should be pointed out that since the step of the parameter of each distortion is uniformly split, the SSIM score difference between adjacent images in each distortion is about $0.01$. 

\begin{figure}[t]	
	\begin{minipage}[t]{0.99\linewidth}
		\centering
		 \includegraphics[width = 0.99\linewidth,trim = {8cm 1cm 9cm 2cm}, clip]{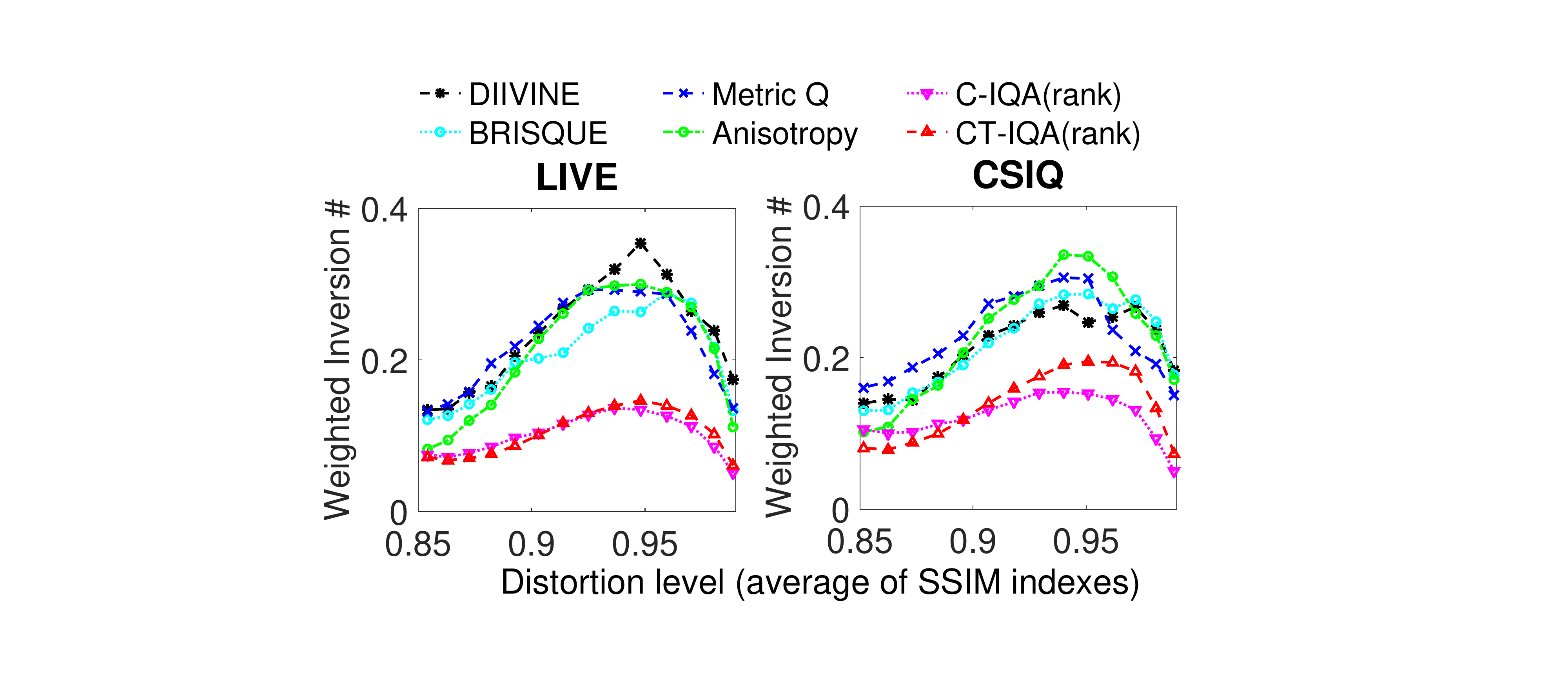}
	 \end{minipage}	 
\caption{Weighted inversion numbers of different NR-IQA methods at 14 distortion levels}
\label{LIVE_OSU_balance}
 \end{figure}

Since C-IQA/CT-IQA are rank-based algorithms, distorted images in one set are sorted by the bubble sort algorithm. The bubble sort algorithm is the same as the traditional one, except that we use C-IQA/CT-IQA to compare each adjacent pair of images in the ranking sequence. In our experiment, the final ranking results are not sensitive to the initial ranking and we start the bubble sort from a random rank. Fig. \ref{LIVE_OSU_balance} shows weighted inversion numbers at 14 distortion levels on two databases. %$x$ axis in Fig. \ref{LIVE_OSU_balance} stands for the average SSIM score of these 8 images and $y$ axis stands for the average Weighted Inversion Number of all the images in the database. 
%(minimum resolution is a distortion-based concept. When two images are distorted in different ways, I don't think there is such phenomenon)

In Table \ref{LIVE_OSU_ability_balance}, we provide the average weighted inversion numbers over 14 distortion levels of different NR-IQA algorithms. It is clear C-IQA and CT-IQA are two of the best NR-IQA algorithms.

 \begin{table*}
	 \centering
	 \caption{Average weighted inversion numbers of different NR-IQA algorithms}
	 \begin{tabular}{|c|c|c|c|c|c|c|} \hline
		  & DIIVINE & BRISQUE & MetricQ & Anisotropy & C-IQA (ranking) & CT-IQA (ranking) \\ \hline	  		 	
		  LIVE  & 0.2328 & 0.2032 & 0.2205 & 0.2064  & \textbf{0.1001} & 0.1026 \\ \hline		 
		  CSIQ &  0.2137 & 0.2168  & 0.2283 & 0.2274 & \textbf{0.1207} & 0.1362 \\ \hline
 \end{tabular} 
	 \label{LIVE_OSU_ability_balance}
 \end{table*}

\subsubsection{Balance ability for bilateral filter}
A series of increasingly denoised images by bilateral filtering, $I_1, I_2,\cdots, I_{30}$, is created for each image the same as the previous part. The SSIM index of the most oversmoothed image $I_{30}$ is between $0.85\pm 0.01$. Because the MSD between the adjacent images are below minimum resolution, Alg. \ref{key_image} is adopted to select the best result. Key images are a set of images among which the MSD is greater than the minimum resolution. Alg. \ref{key_image} first separates the 30 increasingly distorted images into a few parts by key images. Images in the two parts next to the best key image are evaluated based on the two key images on the ends. By doing so, we avoid comparing the adjacent images directly. We set $K_{thresh} = 3.0$ in this experiment.

\begin{algorithm}
\caption{Key Image}\label{key_image}
\begin{algorithmic}

\State{\hspace{-1em}\textbf{Key Images Selection;}}
\State{$key\_img = [1]$}
\State{$key_{num} = 1$}
\For{$i = 1:N$}
\If{$MSE(I_i,I_{pre_{key}(key_{num})})>K_{thresh}$}
	\State{$key\_img = [key\_img,i]$}
	\State{$key_{num} = key_{num}+1$}
\EndIf
\EndFor

%\EndProcedure
\bigskip
\State{\hspace{-1em}\textbf{Key Images Comparision;}}
%\Procedure{\textbf{Textureness Compensation:}}{}
\For{$i = 2:(key_{num}-1)$}
	\If{$CQ(I_{key\_img(i)},I_{key\_img(i-1)})>0 $ and\\
		\quad\quad\quad$CQ(I_{key\_img(i)},I_{key\_img(i+1)})>0$
		}
		\State{$best_{key} = i$}
		\State{\it{break;}}
	\EndIf
	
\EndFor

\bigskip
\State{\hspace{-1em}\textbf{Best Image Selection;}}
\State{$start_{num} = key\_img(best_{key}-1)$}
\State{$end_{num} = key\_img(best_{key}+1)$}
\For{$i=start_{num}:end_{num}$}
	\State{$score_{start}(i) = CQ(I(i),I(start_{num}))$}
	\State{$score_{end}(i) = CQ(I(i),I(end_{num}))$}
\EndFor
\State{$best_{img} = max(score_{start}+score_{end})$}
%\EndProcedure
\end{algorithmic}
\end{algorithm}

The SSIM index difference between the best images chosen by a NR-IQA method and the one chosen by SSIM is used to evaluate NR-IQA methods. There are 59 original images in the two databases and 1770 distorted images in our experiment. 
\begin{figure}[t]
\begin{minipage}[t]{0.48\linewidth}
\centering
\includegraphics[width = 0.99\linewidth,trim = {0cm 0cm 0cm 0cm},clip]{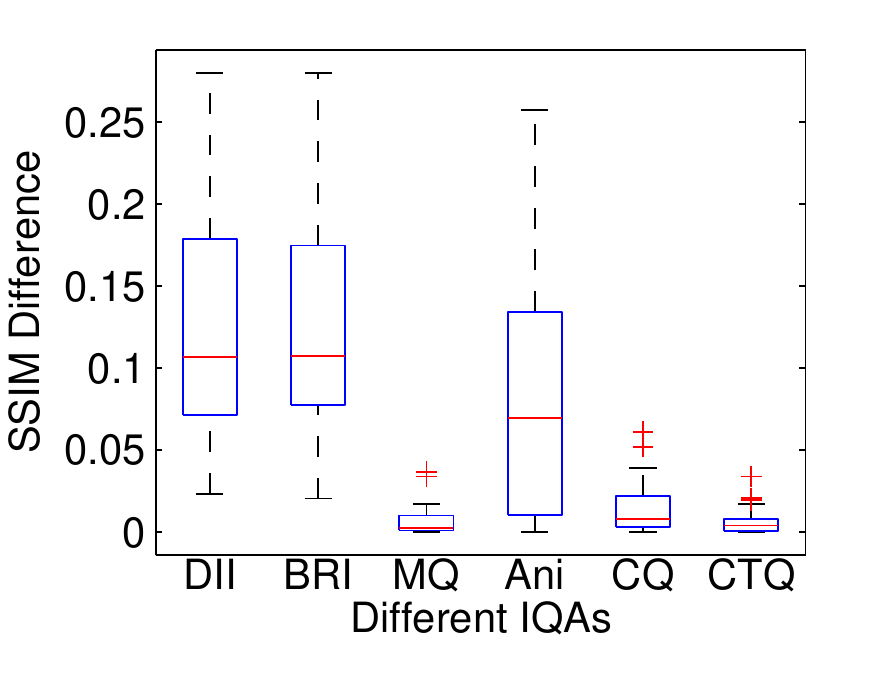}
(a) LIVE
\end{minipage}
\begin{minipage}[t]{0.48\linewidth}
\centering
\includegraphics[width = 0.99\linewidth,trim = {0cm 0cm 0cm 0cm},clip]{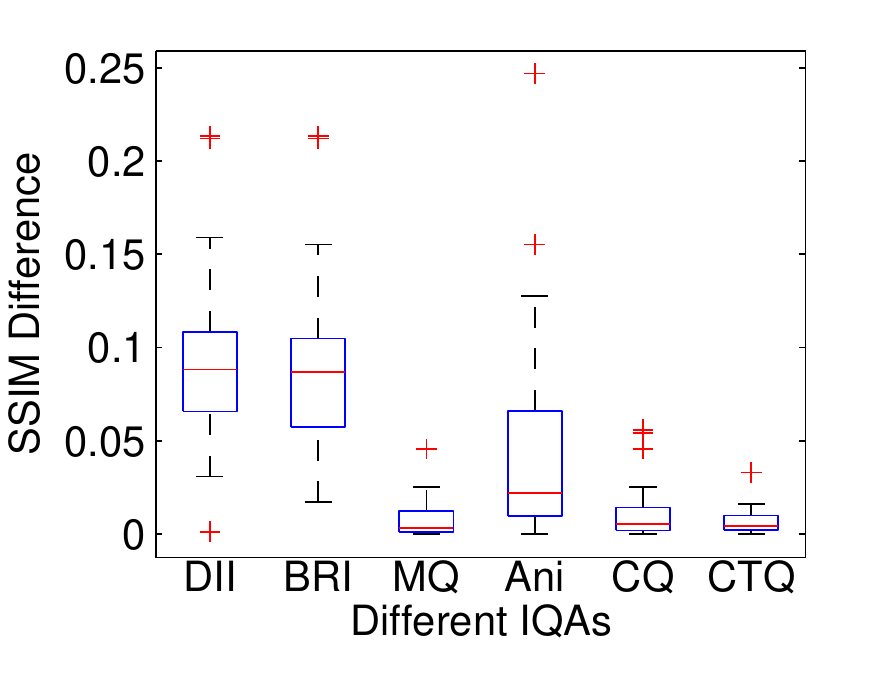}
(b) CSIQ
\end{minipage}
\caption{The SSIM differences between the best images chosen by IQA method and by SSIM from denoised images}
\label{bilateral_database}
\end{figure}
From Fig. \ref{bilateral_database}, we can see that MetricQ and CT-IQA are two of the best NR-IQA algorithms. Table \ref{bilatral_database_table} provides more quantitative evaluations of different NR-IQA algorithms.
\begin{table*}
\centering
\caption{Balancing abilities of blurring and noise on a series of denoised images of different IQAs}
\begin{tabular}{|cc|c|c|c|c|c|c|} \hline
 & & DIIVINE & BRISQUE & MetricQ & Anisotropy & C-IQA & C-T-IQA \\ \hline
\multirow{2}{*}{LIVE}  
	& \multicolumn{1}{ |c| }{median of all SSIM differences} & $1.06\times10^{-1}$ & $1.07\times10^{-1}$ &  $\mathbf{2.36\times10^{-3}}$  & $6.95\times10^{-2}$ &  $1.43\times10^{-2}$ & $3.80\times10^{-3}$  \\ \cline{2-8} 
	& \multicolumn{1}{ |c| }{average of all SSIM differences} & $1.22\times10^{-1}$ &  $1.25\times10^{-1}$   & $6.73\times10^{-3}$ & $8.23\times10^{-2}$  &  $1.43\times10^{-2}$  & {$\mathbf{6.05\times10^{-3}}$}  \\ \cline{2-8} 
	& \multicolumn{1}{ |c| }{average of non-outliers} &  $1.22\times10^{-1}$  & $1.25\times10^{-1}$  & $5.67\times10^{-3}$ & $8.23\times10^{-2}$  & $1.12\times10^{-2}$  & {$\mathbf{3.93\times10^{-3}}$}   \\ \hline
\multirow{2}{*}{CSIQ}  
	& \multicolumn{1}{ |c| }{median of all SSIM differences} & $8.81\times10^{-2}$ & $8.68\times10^{-2}$ & $\mathbf{2.83\times10^{-3}}$ & $2.18\times10^{-2}$ & $5.28\times10^{-3}$ &  $4.05\times10^{-3}$  \\ \cline{2-8} 
	& \multicolumn{1}{ |c| }{average of all SSIM differences} & $9.11\times10^{-2}$ & $8.95\times10^{-2}$ &  $7.65\times10^{-3}$ & $4.30\times10^{-2}$ &  $1.09\times10^{-2}$ & {$\mathbf{6.24\times10^{-3}}$}  \\ \cline{2-8} 
	& \multicolumn{1}{ |c| }{average of non-outliers} &  $8.54\times10^{-2}$ & $8.06\times10^{-2}$ & $6.34\times10^{-3}$ & $3.17\times10^{-2}$ & $6.36\times10^{-3}$ & {$\mathbf{5.32\times10^{-3}}$}       \\ \hline
 \end{tabular} 
\label{bilatral_database_table}
\end{table*}
 
\begin{figure}[t]
\begin{minipage}[t]{0.58\linewidth}
\centering
\includegraphics[width = 0.99\linewidth]{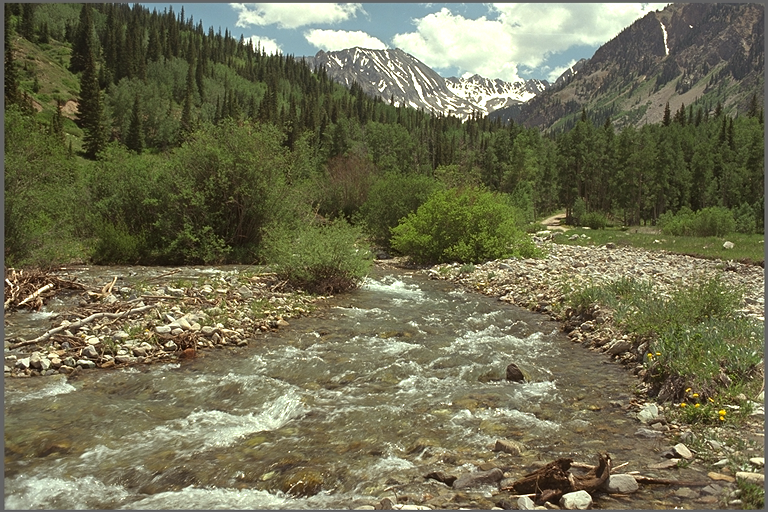}
(a) ``stream'' (from LIVE)
\end{minipage}
\begin{minipage}[t]{0.38\linewidth}
\centering
\includegraphics[width = 0.99\linewidth]{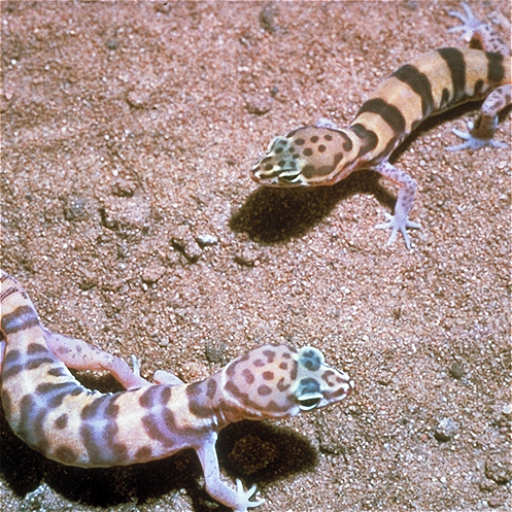}
(b) ``geckos'' (from CSIQ)
\end{minipage}
\caption{Outliers of parameter selection}
\label{outliers}
\end{figure}

In order to have a better understanding of Comparison-based IQA, we show two outliers of C-IQA/CT-IQA in Fig. \ref{outliers}. Without the knowledge of the contents in the scenes, fine textures are regarded as noise in these two images, and both C-IQA/CT-IQA choose oversmoothed images as the best denoised results.

\subsubsection{Balance ability for TV reconstruction}
\label{balancity_recon_sec}
The algorithm used for image reconstruction is introduced in Section \ref{image_reconstruction}. In the experiment, $70\%$ Fourier transform data are used to reconstruct the image and in order to be more realistic, Fourier transform data are distorted by Gaussian noise. The SNR is kept at 20 dB in all reconstruction experiments. All 30 regularization parameter candidates are uniformly selected between $[10^{-5},10^{-1}]$ in logarithmic scale. Because in this experiment, the MSD between adjacent images is above the minimum resolution of C-IQA/CT-IQA, we choose the best image simply by comparing the adjacent images.

\begin{figure}[t]
\begin{minipage}[t]{0.48\linewidth}
\centering
\includegraphics[width = 0.99\linewidth,trim = {0cm 0cm 0cm 0cm},clip]{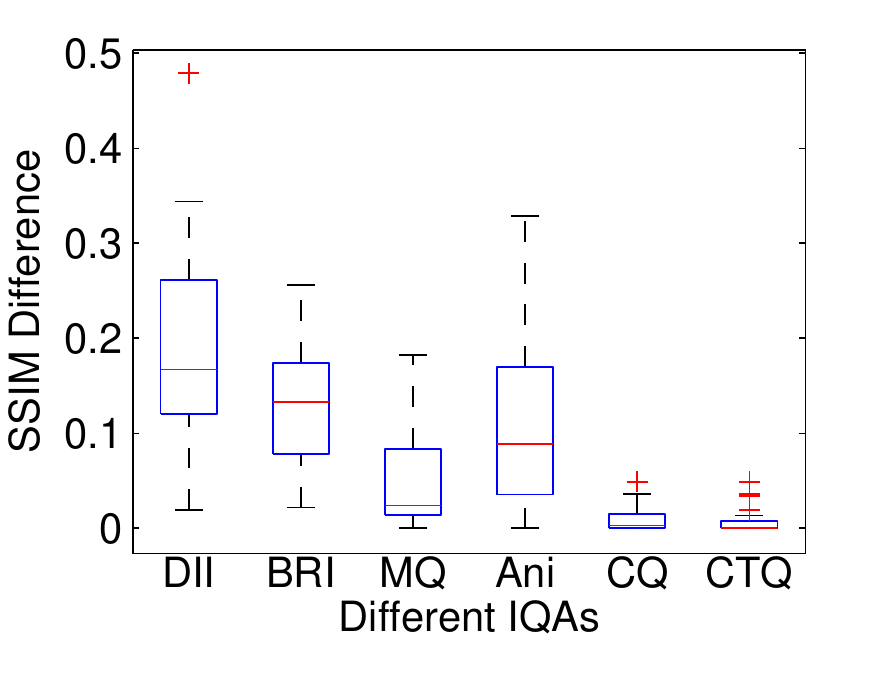}
(a) LIVE
\end{minipage}
\begin{minipage}[t]{0.48\linewidth}
\centering
\includegraphics[width = 0.99\linewidth,trim = {0cm 0cm 0cm 0cm},clip]{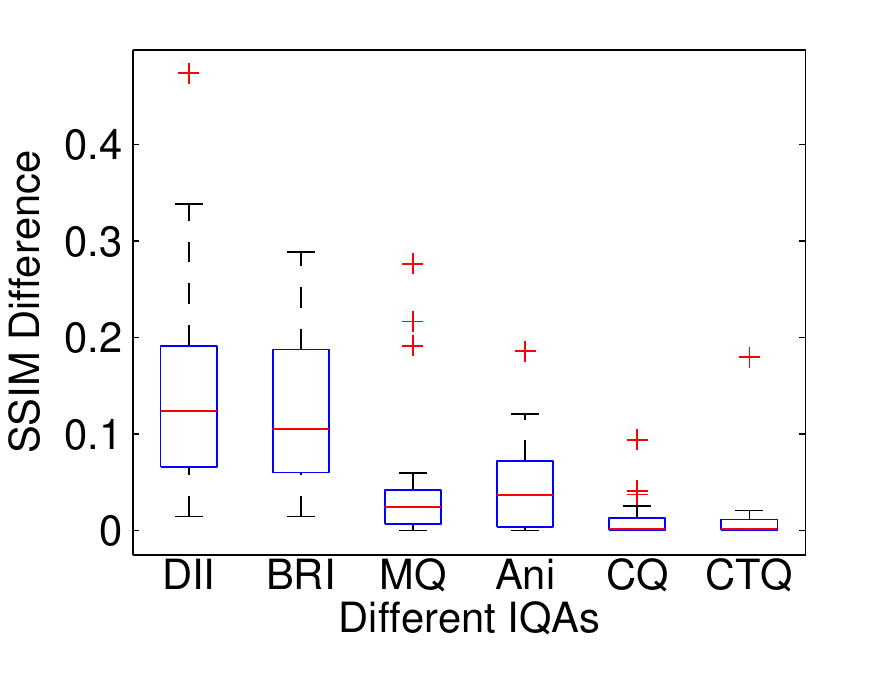}
(b) CSIQ
\end{minipage}
\caption{The SSIM differences between the best images chosen by NR-IQA method and the by SSIM from reconstructed images}
\label{recon_database}
\end{figure}

Th SSIM difference of each IQA algorithm is plotted in Fig. \ref{recon_database}. C-IQA and CT-IQA are the two best IQA algorithms. In Table \ref{recon_database_table}, we provide quantitative evaluation of different NR-IQA algorithms.
The two outliers in TV reconstruction are the same images as shown in Fig. \ref{outliers}.
\begin{table*}
\centering
\caption{Balancing abilities of blurring and noise on a series of reconstructed images of different IQAs}
\begin{tabular}{|cc|c|c|c|c|c|c|} \hline
 & & DIIVINE & BRISQUE & MetricQ & Anisotropy & C-IQA & C-T-IQA \\ \hline
\multirow{2}{*}{LIVE}  
	& \multicolumn{1}{ |c| }{median of all SSIM differences} & $1.67\times10^{-1}$ & $1.33\times10^{-1}$ &  $2.42\times10^{-2}$ & $8.88\times10^{-2}$ &  $2.97\times10^{-3}$ & $\mathbf{0}$  \\ \cline{2-8} 
	& \multicolumn{1}{ |c| }{average of all SSIM difference} & $1.85\times10^{-1}$ &  $1.33\times10^{-1}$   & $5.07\times10^{-2}$ & $1.09\times10^{-1}$  &  $9.92\times10^{-3}$  & {$\mathbf{7.77\times10^{-3}}$}  \\ \cline{2-8} 
	& \multicolumn{1}{ |c| }{average of non-outliers} &  $1.74\times10^{-1}$  & $1.33\times10^{-1}$  & $5.07\times10^{-2}$ & $1.09\times10^{-1}$  & $7.02\times10^{-3}$  & {$\mathbf{2.07\times10^{-3}}$}   \\ \hline
\multirow{2}{*}{CSIQ}  
	& \multicolumn{1}{ |c| }{median of all SSIM difference} & $1.24\times10^{-1}$ & $1.05\times10^{-1}$ & $2.44\times10^{-2}$ & $3.66\times10^{-2}$ & $\mathbf{1.73\times10^{-3}}$ &  $\mathbf{1.73\times10^{-3}}$  \\ \cline{2-8} 
	& \multicolumn{1}{ |c| }{average of all SSIM difference} & $1.43\times10^{-1}$ & $1.23\times10^{-1}$ &  $4.25\times10^{-2}$ & $4.30\times10^{-2}$ &  $\mathbf{1.12\times10^{-2}}$ & {$1.12\times10^{-2}$}  \\ \cline{2-8} 
	& \multicolumn{1}{ |c| }{average of non-outliers} &  $1.31\times10^{-1}$ & $1.23\times10^{-1}$ & $2.19\times10^{-2}$ & $3.81\times10^{-2}$ & $6.02\times10^{-3}$ & {$\mathbf{5.38\times10^{-3}}$}       \\ \hline
 \end{tabular} 
\label{recon_database_table}
\end{table*}

\begin{figure}	
	\begin{minipage}[t]{0.49\linewidth}
		\centering
		\includegraphics[width = 0.99\linewidth]{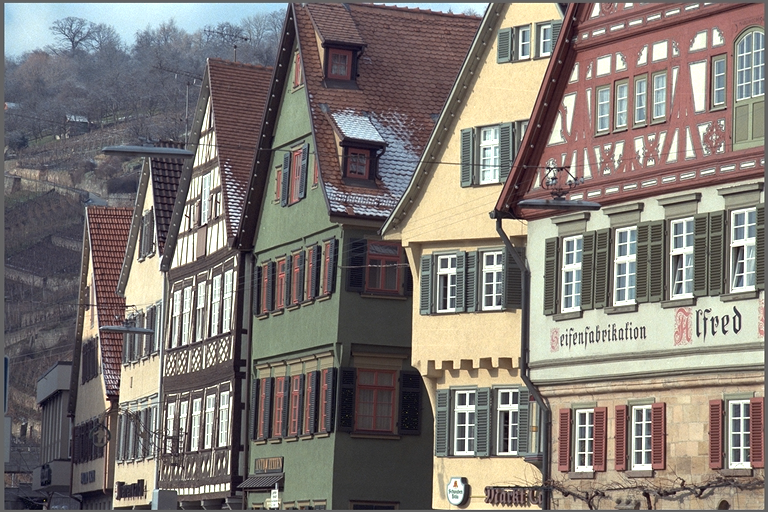}
		(a) ``buildings'' (from LIVE)
	\end{minipage}
	\begin{minipage}[t]{0.49\linewidth}
		\centering
		\includegraphics[width = 0.99\linewidth,trim = {23.5cm 0.5cm 2.5cm 10cm},clip]{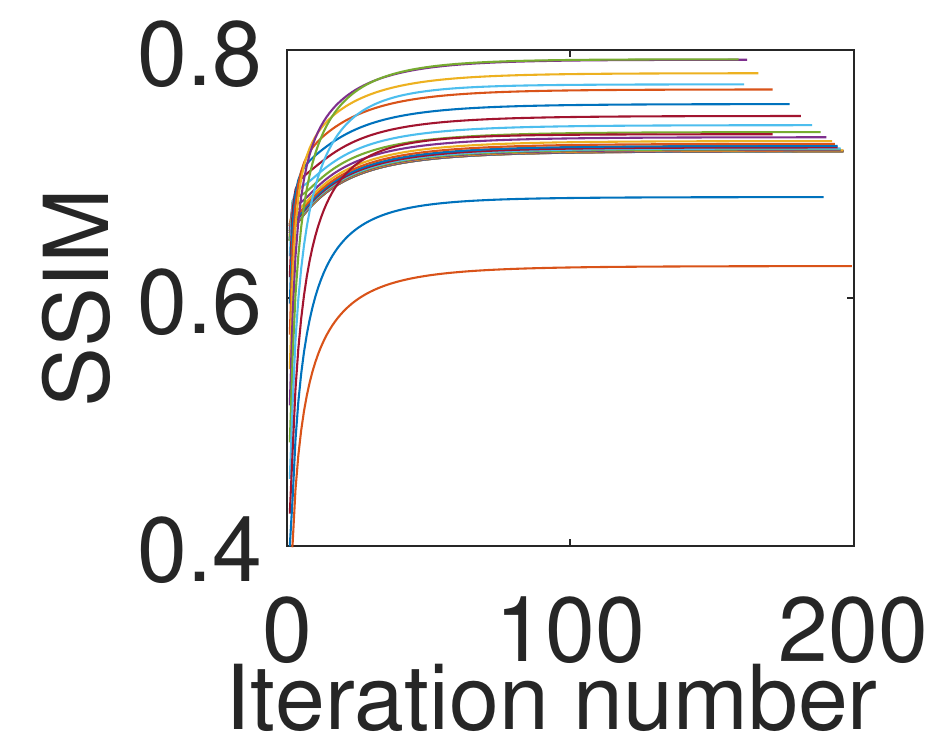}
		(b) Convergence without parameter trimming
	\end{minipage}
	\begin{minipage}[t]{0.49\linewidth}
		\centering
		\includegraphics[width = 0.99\linewidth,trim = {23.5cm 0cm 2.5cm 10cm},clip]{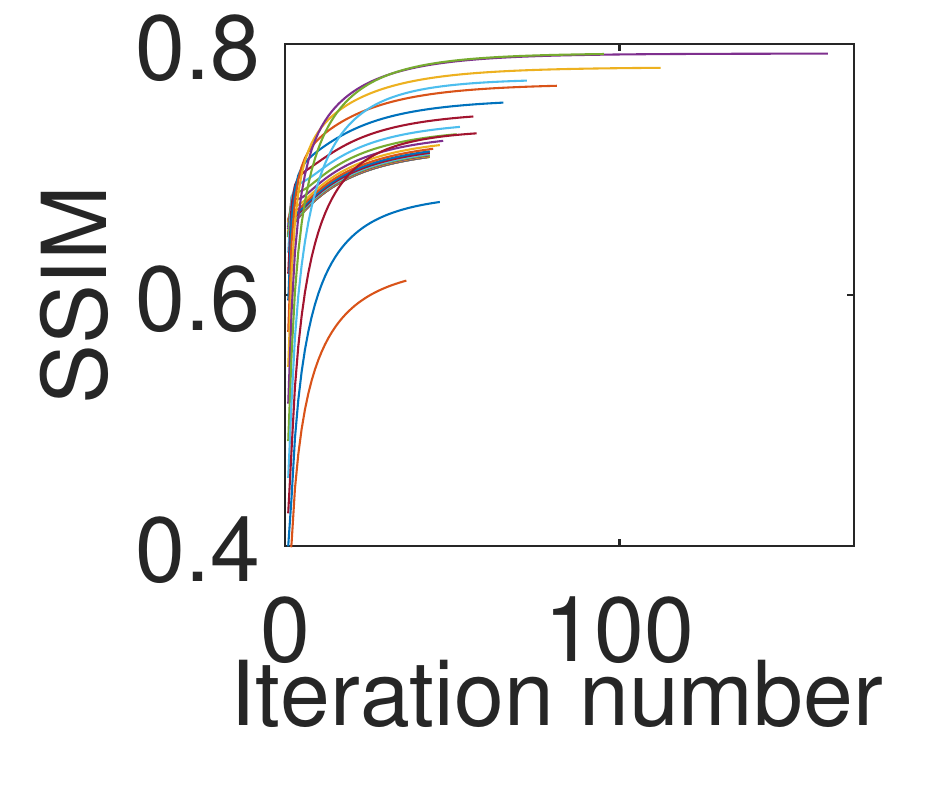}
		(c) Convergence with parameter trimming
	\end{minipage}
	\begin{minipage}[t]{0.49\linewidth}
		\centering
		\includegraphics[width = 0.99\linewidth,trim = {23.5cm 0cm 2.5cm 10cm},clip]{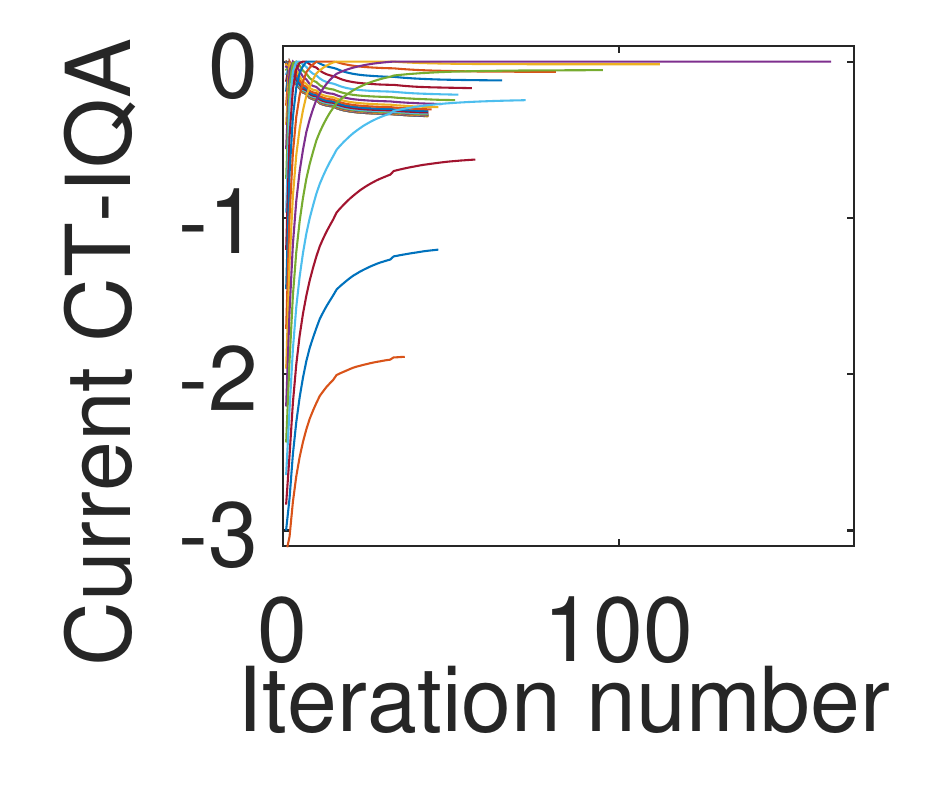}
		(d) Current CT-IQA scores
	\end{minipage}
	\begin{minipage}[t]{0.49\linewidth}
		\centering
		\includegraphics[width = 0.99\linewidth,trim = {23.5cm 0cm 2.5cm 10cm},clip]{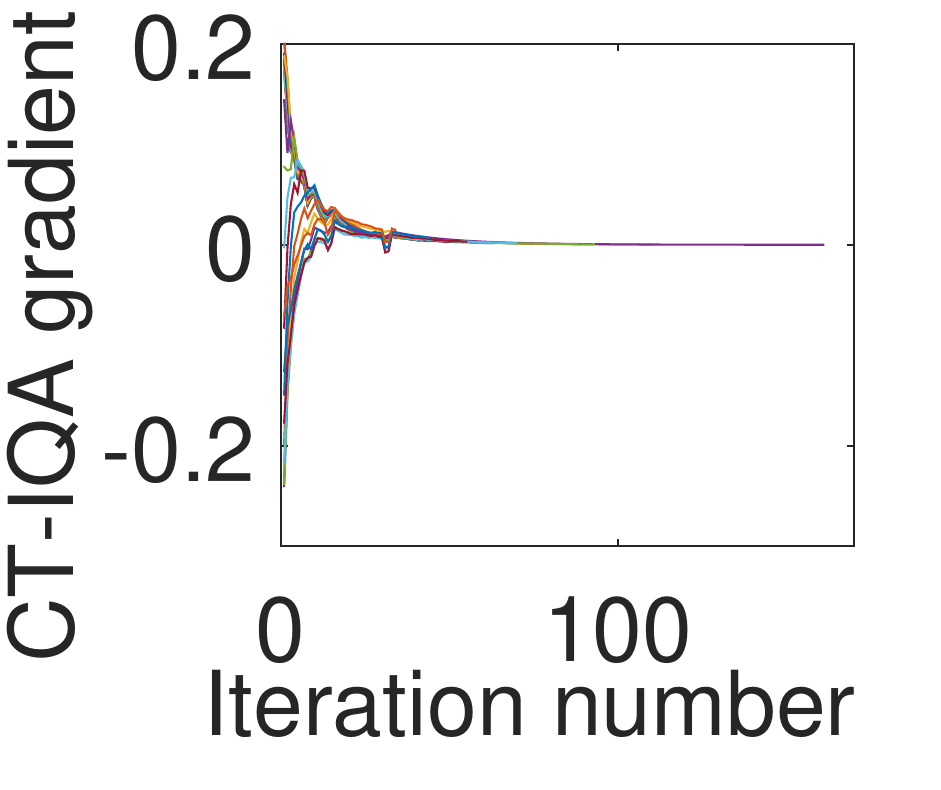}
		(e) Current CT-IQA gradients
	\end{minipage}
	\begin{minipage}[t]{0.49\linewidth}
		\centering
		\includegraphics[width = 0.99\linewidth,trim = {23.5cm 0cm 2.5cm 10cm},clip]{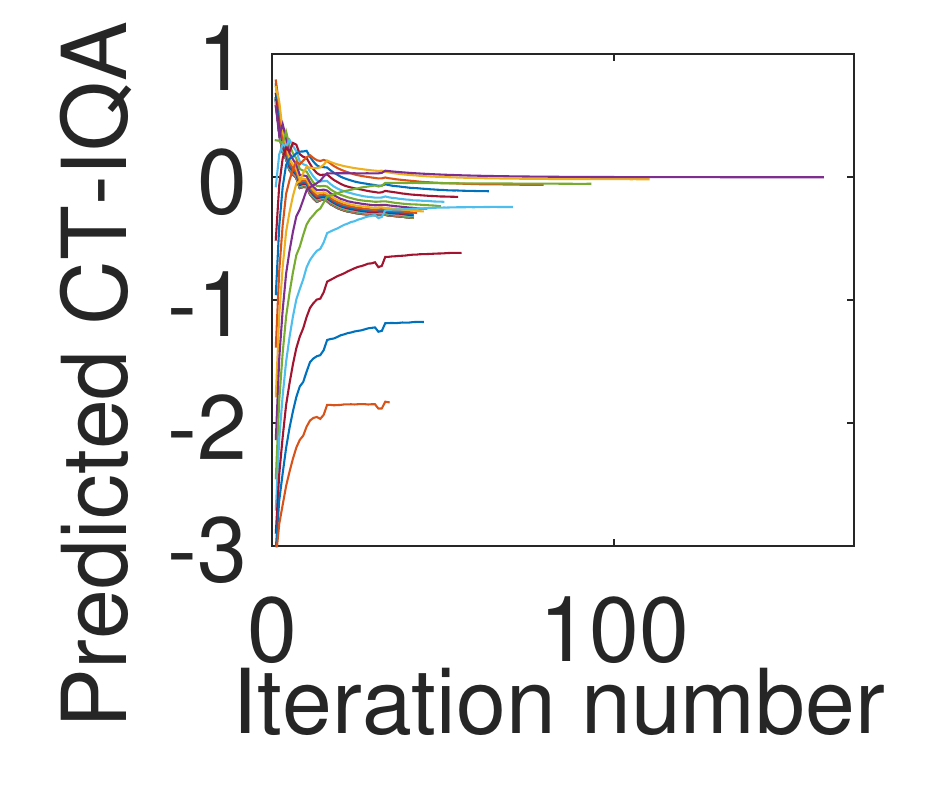}
		(f) Predicted CT-IQA scores
	\end{minipage}
	
	\caption{Comparison between convergence with and without 1-D parameter trimming on ``buildings''}
	\label{trimming_examples}
\end{figure}
\begin{figure}	
	\begin{minipage}[t]{0.25\linewidth}
		\centering
		\includegraphics[width = 0.99\linewidth]{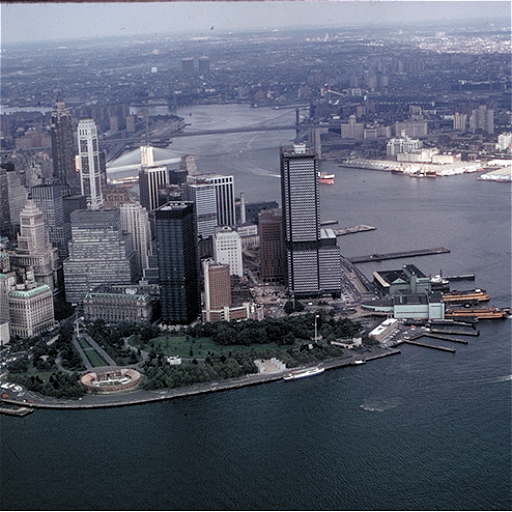}
		(a) ``aerial\_city'' (from CSIQ)
	\end{minipage}
	\begin{minipage}[t]{0.36\linewidth}
		\centering
		\includegraphics[width = 0.99\linewidth,trim = {23.5cm 0cm 3cm 9.5cm},clip]{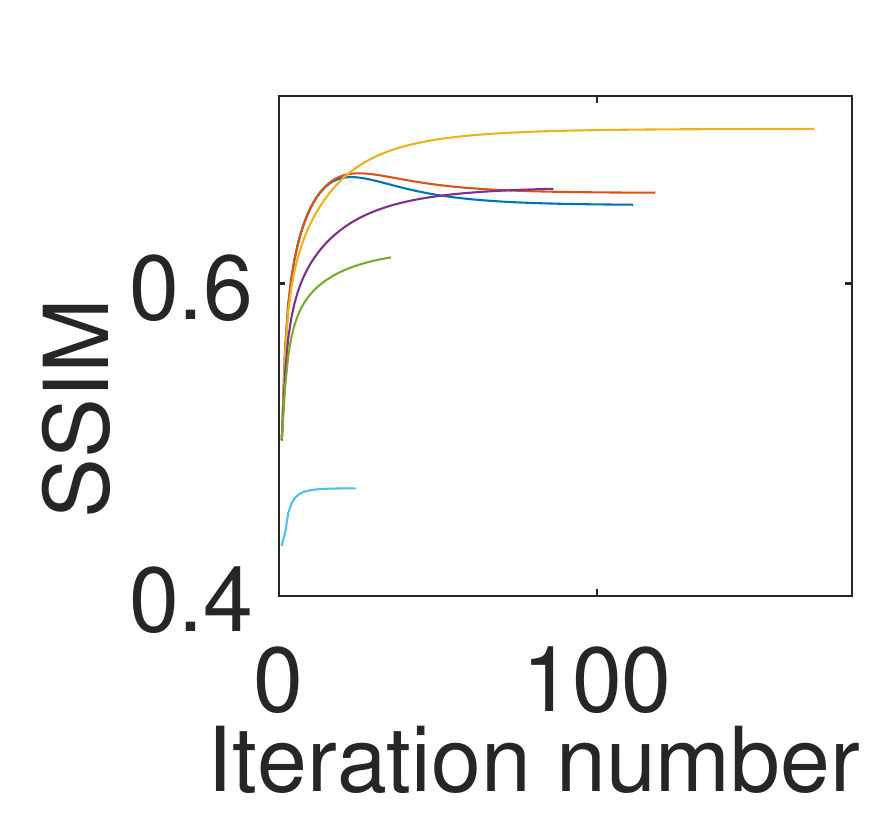}
		(b) Convergence without parameter trimming (different $\gamma$)
	\end{minipage}
	\begin{minipage}[t]{0.36\linewidth}
		\centering
		\includegraphics[width = 0.99\linewidth,trim = {23.5cm 0cm 3cm 9.5cm},clip]{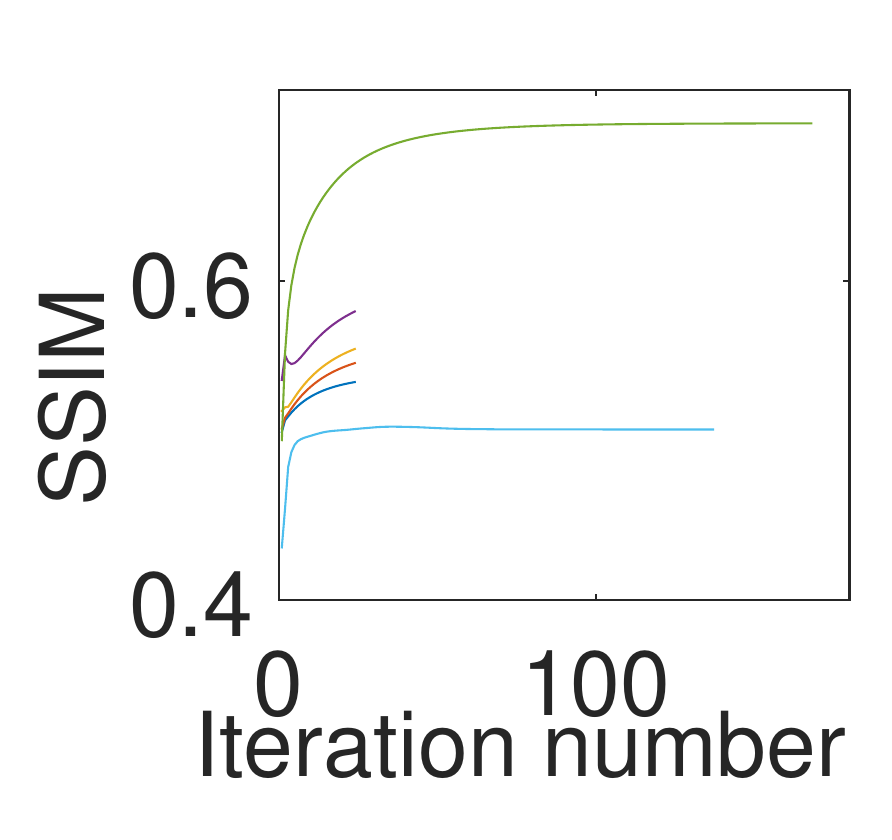}
		(c) Convergence with parameter trimming (different $\beta$)
	\end{minipage}
	
	\caption{Comparison between convergence with and without 2-D parameter trimming on ``aerial city''}
	\label{2Dtrimming_examples}
\end{figure}

\subsection{Application in iterative framework}
In this section, we combine CT-IQA with the parameter trimming framework and show that considerable computation can be saved while preserving the accuracy of parameter selection. %In the one dimension part, we run the reconstruction algorithm introduced in previous part on two databases. In multi-dimension parameters trimming part, we run the reconstruction algorithm with both TV and Haar wavelet regularization term and thus two parameters need to be selected.
\subsubsection{1-D parameter trimming}
In this part, all the parameter settings are the same as the settings in Section \ref{balancity_recon_sec}.
On LIVE\cite{LIVE_dataset}, all the parameters selected with parameter trimming are the same as the parameters selected after convergence; on CSIQ\cite{FR_IQA_review_OSU}, only one of the best parameter selected by parameter trimming is different from the one selected after convergence. Fig. \ref{trimming_examples}(a) is one example image in parameter trimming. In Fig. \ref{trimming_examples}(b) and Fig. \ref{trimming_examples}(c), we show the effectiveness of parameter trimming with SSIM as the quality index (SSIM is only used to demonstrated the trimming process here); Fig. \ref{trimming_examples}(d) -Fig. \ref{trimming_examples}(f) show the changes of three CT-IQA related indices that we actually use to make the trimming decision. From Fig. \ref{trimming_examples}, we can see that the trimming decision achieves the goal of terminating iteration of parameters that is far from the best choice and thus saving computation.

In Table \ref{parameter_trimming_table}, we provide the average numbers of iterations with and without parameter trimming per image on two databases.

\begin{table}
\centering
\caption{Computation saved by parameter trimming}
\begin{tabular}{|p{0.6cm}|p{2.4cm}|p{2.4cm}|p{1cm}|} \hline
 & {Ave \# of iteration without parameter trimming} & Ave \# of iteration with parameter trimming& saved computation (\%)\\  \hline
LIVE& 	4651.9	&		847.7			&	81.78		\\	\hline
CSIQ&		4565.6	&	941.1		&	79.39	\\	\hline
\end{tabular}
\label{parameter_trimming_table}
\end{table}

\subsubsection{2-D parameters trimming}
We include two regularizers, TV and Haar wavelet, for the reconstruction algorithm in this part. Two regularization parameters, $\beta$ and $\gamma$, are for TV and Haar wavelet respectively and both have six parameter candidates. Regularization parameters, $\beta$ and $\gamma$, are uniformly sampled between $[10^{-5},10^{-1}]$ and $[10^{-8},10^{-1}]$ in logarithmic scale. One image from CSIQ\cite{FR_IQA_review_OSU} is tested for 2-D parameter trimming and the best set of parameters is correctly selected by parameter trimming. In Fig. \ref{2Dtrimming_examples}, parameter trimming is illustrated by changing one parameter while fixing the other parameter as the best choice.

\section{Conclusion}
\label{conclusion}
In this paper, we focused on the NR-IQA method and discussed the advantages and drawbacks of two different approaches to the NR-IQA problem, global and local approaches. Inspired by some key concepts put forward in the previous works \cite{anisotropy,MetricQ}, for the first time we designed a comparison-based IQA method and analyzed important properties unique to comparison-based IQA, such as minimum resolution. The novel C-IQA/CT-IQA method includes three primary modules, Content Detection, Contribution and Texture Compensation. At last, the comparison-based IQA compares favorably with other NR-IQA algorithms on two widely used databases\cite{LIVE_dataset, FR_IQA_review_OSU}.

In the experiment (Section \ref{experiments}), we showed that when fine texture with small granularity appears in the image, C-IQA/CT-IQA tends to select the suboptimal result. Integrating the global texture information with local gradient-based structure information is a possible solution to improve the robustness of comparison-based IQA and other NR-IQA algorithms.

We take C-IQA/CT-IQA as a specific implementation of the comparison-based IQA method. By exploiting the extra available information in many image quality assessment applications, other comparison-based IQA methods can be designed for different application scenarios.

\section*{Acknowledgment}
The authors would like to thank Xiang Zhu for providing the MetricQ\cite{MetricQ} code and Anush Krishna Moorthy for explaining the usage of DIIVINE\cite{DIVIINE} code.
\label{sec:ref}
\bibliographystyle{IEEEbib}
\bibliography{refs}
\end{document}